\documentclass{ica}
\usepackage{subfigure}
\usepackage{setspace}
\usepackage{geometry} % added 27-02-2014 Markus Englich
\usepackage{epstopdf}
\usepackage{natbib}
\usepackage[labelsep=period]{caption}  % added 14-04-2016 Markus Englich - Recommendation by Sebastian Brocks
\usepackage{amsmath}
\usepackage{amsfonts}
\usepackage{textcomp}
\usepackage{hyperref}
\usepackage{xcolor}

\usepackage{fancyhdr}

\fancypagestyle{first}{%
    \fancyhead[L]{\includegraphics[height=1.0cm]{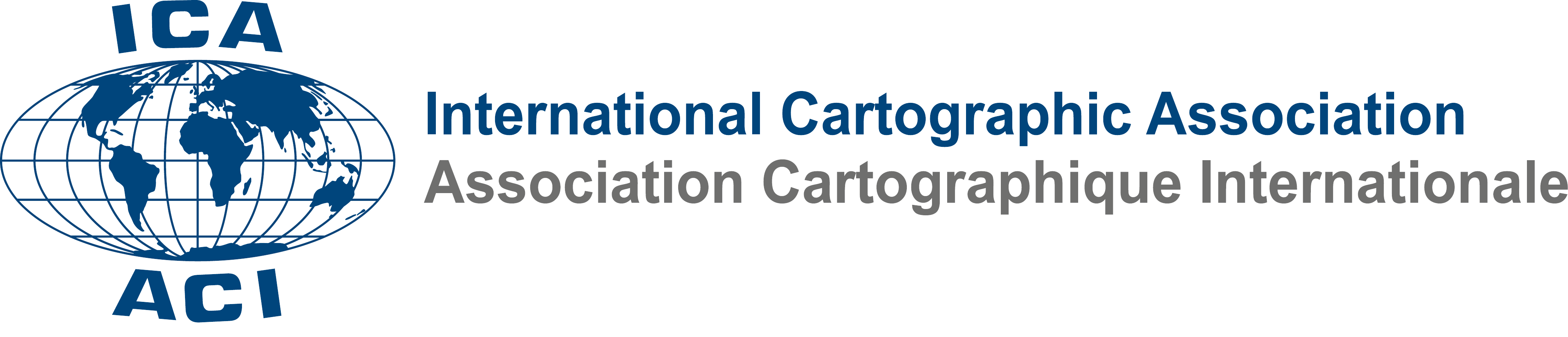}}
    \fancyhead[R]{}
    
}

\begin{document}
\title{Transferring Multiscale Map Styles Using Generative Adversarial Networks}

\author{
 Yuhao Kang\textsuperscript{a},
  Song Gao\textsuperscript{a}\thanks{Corresponding author}, 
  Robert Roth\textsuperscript{b} 
}

\address{
      \textsuperscript{a}Geospatial Data Science Lab, Department of Geography, University of Wisconsin, Madison\\
      \textsuperscript{b}Cartography Lab, Department of Geography, University of Wisconsin, Madison\\
     \textsuperscript{} Email: yuhao.kang@wisc.edu, song.gao@wisc.edu, reroth@wisc.edu\\
}

% If the corresponding author is NOT the final author, always add a % space before the subsequent comma, i.e.
% first author name\textsuperscript{a,}\thanks{Corresponding author} , % second author name \textsuperscript{b}, etc.
% thanks to Niclas Borlin 05-05-2016

\abstract{The advancement of the Artificial Intelligence (AI) technologies makes it possible to learn stylistic design criteria from existing maps or other visual art and transfer these styles to make new digital maps. In this paper, we propose a novel framework using AI for map style transfer applicable across multiple map scales. Specifically, we identify and transfer the stylistic elements from a target group of visual examples, including Google Maps, OpenStreetMap, and artistic paintings, to unstylized GIS vector data through two generative adversarial network (GAN) models. We then train a binary classifier based on a deep convolutional neural network to evaluate whether the \textit{transfer styled map} images preserve the original map design characteristics. Our experiment results show that GANs have great potential for multiscale map style transferring, but many challenges remain requiring future research.
}

%\\ \hline

\maketitle

\thispagestyle{first}

%\saythanks % option for xx
\section{Introduction}\label{intro}
% Sloppy spacing ensures non-overfull lines. Can be removed if this is not an issue.
\sloppy

\begin{figure}
	\centering
	\includegraphics[height=3cm]{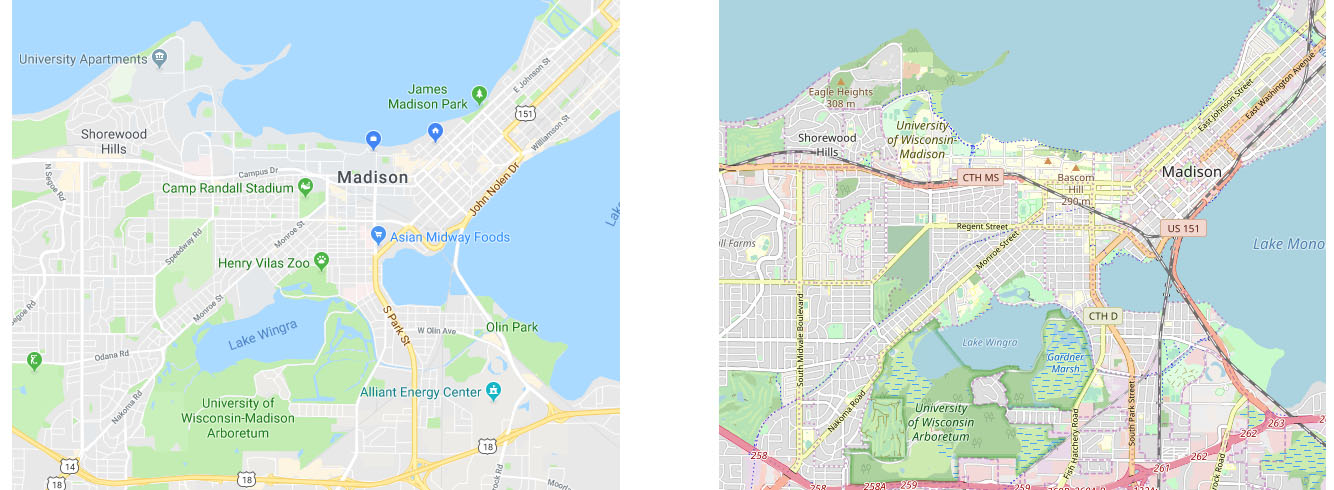}
	\caption{
		The \textit{Google Maps} (left) and \textit{OpenStreetMap} (right) styles for Madison, Wisconsin (USA). \textit{Google Maps} has a flatter visual hierarchy to emphasize labels and points of interests as well as enable vector overlays, whereas \textit{OpenStreetMap} is more visually complex and includes a wider variety of features and symbols. }
	\label{fig:osm_google_madison}
\end{figure}

%\textcolor{red}{
A \textit{map style} is an aesthetically cohesive and distinct set of cartographic design characteristics \citep{kent2009stylistic}. The map style sets the aesthetic tone of the map, evoking a visceral, emotional reaction from the audience based on the interplay of form, color, type, and texture \citep{gao2017designing,rothforth}. Two maps can have a very different look and feel based on their map style, even if depicting the same information or region (Figure \ref{fig:osm_google_madison}; see \cite{stoter2005generalisation,kent2009stylistic} for comparisons of in-house styles of national mapping agencies). Arguably, map styling—and the myriad design decisions therein—is a primary way that the cartographer exercises agency, authorship, and subjectivity during the mapping process (see \cite{CPcp73-buckley-jenny} for recent discussions on aesthetics, style, and taste).
%}

%\textcolor{red}{
Increasingly, web cartographers need to develop a coherent and distinct map style that works consistently across multiple map scales to enable interactive panning and zooming of a “map of everywhere” \citep{roth2011typology}. Such multiscale map styling taps into a rich body of research on generalization and multiple representation databases in cartography (see \cite{mackaness2011generalisation} for a compendium developed by the ICA Commission on Generalization). A large number of generalization taxonomies now exist to inform the multiscale map design process (e.g., \cite{delucia1987comprehensive,christophe2016map, mcmaster1992generalization,regnauld2007synoptic,foerster2007towards,stanislawski2014generalisation,rap,shen2018new}), most of which focus on vector geometry operations for meaningfully removing detail in geographic information (e.g., simplify, smooth, aggregate, collapse, merge).
%}

%\textcolor{red}{
\cite{brewer2007framing} argue that adjusting the symbol styling can have as great an impact in the legibility of multiscale map designs as other selection or geometry manipulations. Accordingly, \cite{roth2011typology} discuss how cartographers can manipulate the \textit{visual variables}, or fundamental building blocks of graphic symbols (e.g., shape, size, orientation, dimensions of color like hue, value, saturation, and transparency), to promote legibility and maintain a coherent style across map scales. A number of web mapping services and technologies now exist to develop and render such multiscale map style rules as interlocking \textit{tilesets}, such as \textit{CartoCSS}\footnote{\url{https://carto.com/developers/styling/cartocss/}}, \textit{Mapbox Studio}\footnote{\url{https://www.mapbox.com/designer-maps/}},\textit{TileMill}\footnote{\url{https://tilemill-project.github.io/tilemill/}}, or \textit{TileCache}\footnote{\url{http://tilecache.org/}}.
Beyond authoritative or classic map styles (see \cite{muehlenhaus2012if} for a review), these tools enable multiscale web map styling that is exploratory, playful, and even subversive (for instance, see \cite{christophe2012expressive} for examples of multiscale map styling using Pop Art as inspiration). Despite these advances, establishing a map style that works across regions and scales remains a fundamental challenge for web cartography, given the wide array of stylistic choices available to the cartographer and the limited guidance for integrating creative, artistic styles into multiscale maps like \textit{Google Maps}\footnote{\url{https://www.google.com/maps}} and \textit{OpenStreetMap} (OSM)\footnote{\url{https://www.openstreetmap.org}}.
%}

%\textcolor{red}{
Here, we ask if \textit{artificial intelligence} (AI) can help illuminate, transfer, and ultimately improve multiscale map styling for cartography, automating some of the multiscale map style recreation and assisting the cartographer in developing novel representations. Our work draws from the active symbolism paradigm  in cartography and visualization \citep{armstrong2018retrospective}, in which “the production of maps switches from a sequence of actions taken by a mapmaker to a process of specifying criteria that are used to create maps using intelligent agents”. Specifically, whether AI can \textit{learn} map design criteria from existing map examples (or visual art) and then \textit{transfer} these criteria to new multiscale map designs.
%}

%\textcolor{red}{
Latest AI technology advancements in the past decade include a range of deep learning methods developed primarily in computer science for image classification, segmentation, objection localization, style transfer, natural language processing, and so forth \citep{lecun2015deep,goodfellow2016deep,gatys2016image}. Recently, GIScientists and cartographers, along with computer scientists have been investigating various AI and deep learning applications such as geographic knowledge discovery \citep{mao2017geoai,hu2018geoai},  map-type classification \citep{zhou2018deep}, scene classification \citep{zou2015deep,srivastava2018multilabel,law2018street,zhang2018measuring,zhang2019social}, scene generation \citep{deng2018like}, automated terrain feature identification from remote sensing imagery \citep{li2018automated}, automatic alignment of geographic features in contemporary vector data and historical maps \citep{duan2017automatic}, satellite imagery spoofing \citep{xu2018satellite}, spatial interpolation \citep{zhu2019spatial}, and environmental epidemiology \citep{vopham2018emerging}. Relevant to our work on multiscale map style, a new class of AI algorithms called \textit{generative adversarial networks} (GANs) have been developed to generate synthetic photographs that mimic real ones \citep{goodfellow2014generative}. The GANs input real photographs to train the model, and the resulting output photographs look at least superficially authentic to human observers, suggesting a potential application for the multiscale map styling. Several promising studies have used GANs combined with multi-layer neural networks to transfer the styles of existing satellite imagery and vector street maps \citep{isola2017image, CycleGAN2017,xu2018satellite, ganguli2019geogan}. However, several research questions and uncertainty concerns remain. First, which feature types, symbol styling, and zoom levels best enable map style transfer? Second, which AI algorithm or combinations of algorithms work best for map style transfer? Finally, how usable are the resulting maps after style transfer; do the results appear as authentic maps or not?  
%}

%\textcolor{red}{
To this end, we propose a novel framework to transfer existing style criteria to new multiscale maps using GANs without the input of CartoCSS map style configuration sheets. Specifically, the GANs learn (1) which visual variables encode (2) which map features and distributions at (3) which zoom levels, and then replicate the style using the most salient combinations.  In order to evaluate the results of our framework, we then train a deep convolutional neural network (CNN) classifier to judge whether the outputs with transferred map styling still preserve the input map characteristics.
%}

The paper proceeds with four additional sections. In Section \ref{sec:methodology}, we describe the methods framework, including data collection and preprocessing, tiled map generation, and the paired and unpaired GAN models based on \textit{Pix2Pix} and \textit{CycleGAN}  respectively. We then describe in Section \ref{sec:experiment} an experiment using geospatial features in Los Angeles and San Francisco (USA) to test the feasibility and accuracy of our framework. Specifically, we provide both a qualitative visual assessment and quantitative assessment of two different GAN models, \textit{Pix2Pix} and \textit{CycleGAN}, at two different map scales.  We discuss potential applications with challenges in Section \ref{sec:discussion} and offer conclusions and future work in Section \ref{sec:conclusion}.

\section{Methods}\label{sec:methodology}
\subsection{Overview} \label{sec:framework}
\begin{figure}
	\centering
	\includegraphics[height=0.48\linewidth]{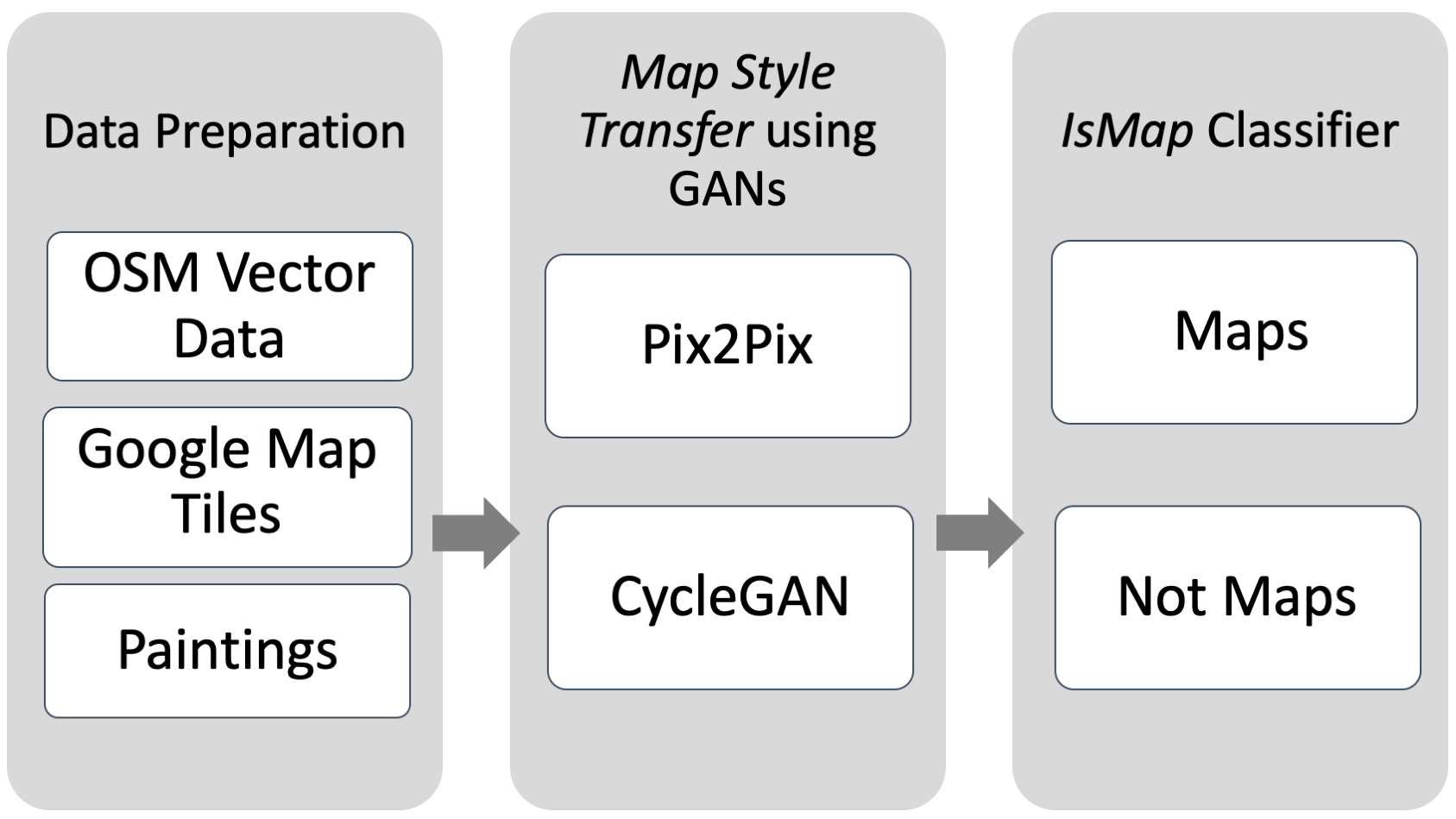}
	\caption{The methodology framework for map style transfer and evaluation: (1) data preparation, (2) map style transfer using GANs, (3) \textit{IsMap} classifier.}
	\label{fig:framework}
\end{figure}

Our proposed methods framework includes three stages as shown in Figure \ref{fig:framework}. First, we prepare unstyled or “raw” GIS vector data from a geospatial data source that we wish to style (here \textit{OpenStreetMap} (OSM) vector data, which is given an initial simple styling for the purpose of visual display; details below) as well as example styled data sources we wish to reproduce and transfer (here \textit{Google Maps} tiles and painted visual art). Second, we configure two generative adversarial network methods to learn the multiscale map styling criteria: \textit{Pix2Pix}, which uses paired training data between the target and example data sources, and \textit{CycleGAN}, which can use unpaired training data (details below). Third, we employ a \textit{deep convolutional neural network} (CNN) classifier (described as \textit{IsMap} below) to judge whether the outputs with transferred map styling do or do not preserve map characteristics  \citep{krizhevsky2012imagenet,evans2017livemaps}.

%As shown in Figure \ref{fig:framework}, there are three stages in our proposed framework of style transfer for maps. 
%First, we need to prepare no-styling raw GIS vector data from one geospatial data source, which will be rendered with a simple style after data preprocessing (More details in the following).
%Another dataset that needs to be prepared are the target-styled maps (i.e., Google Maps tiles and painting-style pictures in this research). Second, two generative adversarial network methods, namely \textit{Pix2Pix} which is based on paired training data, and \textit{CycleGAN} which is based on unpaired training data are used to generate transferred-styled maps from simple-style tiled maps into the target styles automatically.
%After that, in order to evaluate the results, a CNN classifier (IsMap) is then trained to judge whether the outputs with transferred map-styling still preserve map characteristics or not.

\subsection{Data Preparation and Preprocessing}

%\textcolor{red}{
	Our framework requires two types of map layers as inputs: we refer to these as \textit{simple styled maps} and \textit{target styled maps}. We generate the \textit{transfer styled maps} by incorporating the geographic features from \textit{simple styled maps} with the aesthetic styles of the \textit{target styled maps}. We then collect and generate the input map layers as raster web map tilesets that comprise square images, but contain styling symbols that represent different types of geographic features (e.g., buildings, lakes, roads, and so on).	
%Our workflow requires two types of map layers as inputs: we refer to these as \textit{simple-styled maps} and \textit{target-styled maps}. We generate the transferred style by incorporating the geographic features from \textit{simple-styled maps} with the aesthetic styles of the \textit{target-styled} maps. We then collected and generate the input map layers as raster web map tilesets that look like square images, but contain styling symbols that represent different types of geographic features (e.g., buildings, lakes, roads, and so on).
%}

%\textcolor{red}{
Tiled map services are among the most popular web mapping technologies for representing geographical information at multiple scales \citep{CPcp78-roth-et-al}. Such web map tilesets interlock using a multi-resolution, hierarchical pyramid model. Within this pyramid model, map scale is referred to as \textit{zoom level} and expressed in 1-20 notation, with 1 denoting the smallest cartographic scale (i.e., zoomed out) and 20 the largest cartographic scale (i.e., zoomed in). While the spatial resolution gets coarser from the top to the bottom of the tile pyramid, the size of each image tile in the tileset remains across zoom levels \citep{peterson2011travels,shook2014mapping}, typically captured at 256x256 pixels (8-bit). Therefore, serving pre-rendered image tiles typically is less computationally demanding than dynamically rendering vector map tiles in the browser. We used two popular tilesets for this study: \textit{OpenStreetMap}, 
which we downloaded in raw vector format without any map styles and served as a tiled web service for the \textit{simple styled maps} case using \textit{GeoServer\footnote{\url{http://geoserver.org/}}}, and \textit{Google Maps}, which we acquired using their API as the \textit{target styled maps} case. Within \textit{OSM} data, multiple classes of features exist for each geometry type (i.e., point, line, and polygon). For the \textit{simple styled maps}, we rendered these different classes using different colors and subtle transparency so that they could be visually discriminated in the resulting tileset. We used the \textit{Spherical Mercator} (EPSG:900913) coordinate system for georeferencing the \textit{OSM} tiles to ensure they aligned with the \textit{Google Maps} tileset.

\subsection{GANs}
%Next, we utilized the generative adversarial networks (GANs) to generate \textit{transfer styled map} images by combining the geographic features of the \textit{simple styled maps} and the learned map style from the \textit{target styled maps}. GANs have two primary components
 %\citep{goodfellow2014generative}: the generator $G$, which generates fake outputs that mimic real examples using the upsampling vectors of random noise, and the discriminator $D$, which distinguishes the real and fake images according to the downsampling procedure. Following the format of an adversarial loss, $G$ is optimized when the visual features of the reproduced transfer images have a similar distribution with the ground truth target style and the fake images generated by $G$ cannot be distinguished by the discriminator $D$. The training procedures of both $G$ and $D$ occur simultaneously. Overall, the objective function of the procedure is to minimize the maximum loss as:

Next, we utilized the GANs to generate \textit{transfer styled map} images by combining the geographic features of the \textit{simple styled maps} and the learned map style from the \textit{target styled maps}. GANs have two primary components \citep{goodfellow2014generative}: the generator $G$, which generates fake outputs that mimic real examples using the upsampling vectors of random noise, and the discriminator $D$, which distinguishes the real and fake images according to the downsampling procedure. Following the format of an adversarial loss, $G$ iterates through a present number of epochs (an entire dataset is passed forward and backward in one
epoch through the deep learning neural network) and becomes optimized when the visual features of the reproduced transfer images have a similar distribution with the ground truth target style and the fake images generated by $G$ cannot be distinguished by the discriminator $D$. The training procedures of both $G$ and $D$ occur simultaneously.

%Then, we utilized the generative adversarial networks (GANs) to generate transferred-styled map images by combining the geographic features of the simple-styled maps and a specified map style from the target-styled maps. There are two components in GANs \cite{goodfellow2014generative}. One is called generator $G$, which generates fake outputs that look like the real ones from the upsampling vectors of random noise.
%The other is called discriminator $D$, which aims at distinguishing the real and fake images according to the downsampling procedure. The key success of the GANs is an adversarial loss which can learn the mapping through the input imagery data and the translated outputs. In other words, $G$ is optimized so that the visual features of reproduced images have the similar distribution with the ground-truth and the generated fake images by $G$ can not be distinguished by the discriminator $D$. The training procedures of both $G$ and $D$ occur at the same time.  Overall, the objective function of the procedure is to minimize the maximum loss as follows:
\begin{equation}
      \begin{aligned}
            \operatorname*{min}_G
            \operatorname*{max}_D
            V(D, G)=\mathbb{E}_{x{\sim}p_{data}(x)}[logD(x)]+ \\
            \mathbb{E}_{z{\sim}p_{z}(z)}[log(1-D(G(z))] \\
      \end{aligned}
\end{equation}
where \(x\) is a real image and \(z\) is the random noise. 

Since the original GAN aims at generating fake images that have a similar distribution of features in the entire training dataset, it may not be suitable for generating specific types of images under certain conditions. Therefore, \cite{mirza2014conditional} proposed the Conditional GAN (C-GAN) with auxiliary information to generate images with specific information. Different from the original GAN, the C-GAN adds hidden layers \(y\) that contain extra conditional information in generator $G$ and discriminator $D$. The objective function is as:
%Since the original GAN  aims at generating fake images that have very similar distribution of features to the whole training dataset, it may not be suitable for generating specific types of images under certain conditions. Therefore, the Conditional GAN (C-GAN) was proposed with auxiliary information to generate images with specific information \cite{mirza2014conditional}. Different from the GAN, the C-GAN adds hidden layers \(y\) which contain extra conditional information in generator $G$. And in the discriminator $D$, \(y\) is also represented as input.
%The objective function is as follows:
\begin{equation}
      \begin{aligned}
            \operatorname*{min}_G
            \operatorname*{max}_D
            V(D, G)=\mathbb{E}_{x{\sim}p_{data}(x)}[logD(x|y)]+ \\
            \mathbb{E}_{z{\sim}p_{z}(z)}[log(1-D(G(z|y))] \\
      \end{aligned}
\end{equation}

The auxiliary information in the C-GAN can take many forms of input, such as categorical labels that generate images in a specific category (e.g., food, railways; \cite{mirza2014conditional}), and embedded text to generate images from annotations \citep{reed2016generative}. 
For multiscale map styling, the \textit{target styled maps} represent auxiliary information, making the C-GAN more suitable for our research. 

There are two popular types of C-GAN: paired and unpaired. Paired C-GAN uses image-to-image translation to train a model on two paired groups of images, with output combining content from one image and the style from the other image. Unpaired C-GAN also completes an image-to-image translation, but with the transfer of images between two related domains \(X\) and \(Y\) in the absence of paired training examples. In this research, we tested both methods, using the \textit{Pix2Pix} and the \textit{CycleGAN} respectively, to examine their suitability for multiscale map style transfer.

\subsection{Pix2Pix}
\textit{Pix2Pix} is a paired C-GAN algorithm that learns the relationship between the input images and the output images based on the paired-image training set \citep{isola2017image}. In addition to minimizing the objective loss function of general C-GAN, the \textit{Pix2Pix} generator trains not just to fool the discriminator, but also to produce ground truth-like output. The objective function of the extra generator is defined as:
%We use the Pix2Pix framework that is developed by \cite{isola2017image} to generate transferred-styled maps with target map styles from the training dataset.
%Pix2Pix is an image-to-image translation algorithm that can learn the mapping between the input images and the output images based on the paired-image training set. In addition to minimizing the objective loss function of  general C-GAN, the generator is also trained to produce near ground-truth output, while it is not just to fool the discriminator.  And the objective function of the extra generator is defined as follows:
\begin{equation}
     \mathcal{L} _{L1}(G)= \mathbb{E}_{x,y,z}[\parallel y-G(x,z)\parallel_1]
\end{equation}
By combing the two objective functions, the final objective function is computed as:
\begin{equation}
      \begin{aligned}
            \mathcal{L}_{Pix2Pix}=
            \mathcal{L}_{cGAN}(G, D)+\lambda \mathcal{L}_{L1}(G)
      \end{aligned}
\end{equation}
For this research, we paired the \textit{OpenStreetMap} and \textit{Google Maps} tiles for the same locations and at the same zoom levels as the input dataset for training the \textit{Pix2Pix} model.

\subsection{CycleGAN}
%As aforementioned that \textit{Pix2Pix} requires the paired-training samples. However, map datasets and a target artistic style (such as the Monet painting style) cannot be paired. In addition, we also want to extract the map style from one map and convert it to another map even they might have different geographic extents.
%Therefore, another model CycleGAN is also tested in our research. Compared with Pix2Pix, CycleGAN aims at learning the mapping from a source domain to a target domain \cite{CycleGAN2017}. In other word, there is no need to input images with the same geographic view but just need to input two groups of training datasets that have different styles respectively. To achieve this, there are two mappings in this model: 

\textit{Pix2Pix} is appropriate for pairing two map tilesets containing the same geographic extents and scales. However, \textit{Pix2Pix} cannot transfer a target artistic style from non-map examples (e.g., a Monet painting) to a map tileset. Compared with \textit{Pix2Pix},	\textit{CycleGAN} learns the style from one specific source domain (different styles of images) and transfer to a target domain \citep{CycleGAN2017}. In other words, \textit{CycleGAN} does not require two input images with the same geographic extent, but instead just two input training datasets that have different visual styles. \textit{CycleGAN} establishes two associations  to achieve the style transfer: 
\(G: X\)$\,\to\,$\(Y\) and 
\(F: Y\)$\,\to\,$\(X\).
Two adversarial discriminators \(D_X\) and \(D_Y\) are trained respectively, where \(D_X\) distinguishes the images in dataset \(X\) and images generated by \(F(y)\), and \(D_Y\) distinguishes the images in dataset \(Y\) and images generated by \(G(x)\).
The objective function for establishing the relationship of the images in domain \(X\) to domain \(Y\) is represented as:
\begin{equation}
      \begin{aligned}
            \mathcal{L}_{GAN}(G, D_Y, X, Y)=\mathbb{E}_{y\sim p_{data}(y)}[logD_Y(y)]\\
            +\mathbb{E}_{x\sim p_{data}(x)}[log(1-D_Y(G(x))]
      \end{aligned}
\end{equation}
A similar adversarial loss for transforming the images in domain \(Y\) to another domain \(X\) is introduced as \(\mathcal{L}_{GAN}(F, D_X, Y, X)\),
where \(G\) generates images that look similar to the images in the other domain, and \(D\) distinguishes the fake images and the real images.
\textit{CycleGAN} also introduces an extra loss called the cycle consistency loss.
After \(G\) and \(F\) generate images with the similar distribution to the input domain, the cycle consistency loss guarantees when input the images generated to the other generator, the generated images can be restored to the original domain.
In other words, \(x \to G(x) \to F(G(x)) \approx x \). More details can be found in \citet{CycleGAN2017}.
The cycle consistency loss is expressed as:
\begin{equation}
      \begin{aligned}
            \mathcal{L}_{cyc}(G, F)=\mathbb{E}_{x\sim p_{data}(x)}[\parallel F(G(x))-x\parallel_1]\\
            +\mathbb{E}_{y\sim p_{data}(y)}[\parallel G(F(y))-y\parallel_1]
      \end{aligned}
\end{equation}

By combining the two adversarial losses and the cycle consistency loss, the full objective function is expressed as:
\begin{equation}
      \begin{aligned}
            \mathcal{L}_{CycleGAN}=\mathcal{L}_{GAN}(G, D_Y, X, Y)\\
            +\mathcal{L}_{GAN}(F, D_X, Y, X)\\
            +\lambda\mathcal{L}_{cyc}(G,F).
      \end{aligned}
\end{equation}

\subsection{\textit{IsMap} Classifier}
Again, GANs' success in map style transfer relies on the adversarial loss forcing the generated maps to be indistinguishable from the input \textit{target-styled maps}. In addition to the loss curve reported in the model training process, we employ a deep CNN-based binary classifier called \textit{IsMap} to judge whether the \textit{transfer styled maps} are perceived as maps \citep{evans2017livemaps}. CNNs can produce significant improvements in image classification tasks compared with other machine learning models \citep{huang2017densely,maggiori2017convolutional}.
However, the deeper the neural network, the greater the computational costs. Based on existing literature review and comparison on ImageNet, we chose the \textit{GoogleNet/Inception-v3} deep neural network model. More details about the GoogleNet/Inception architecture is available in \cite{szegedy2016rethinking}.

%Since we just need to make a binary classification of images about whether the inputs are maps or not, we used the original deep CNN architecture that is proposed by \cite{krizhevsky2012imagenet}. It contains eight layers in total. The first five are convolutional layers and some of them followed by max pooling layer (which is a sample-based discretization process), and the last three are fully connected layers.

\begin{figure}
	\centering
	\includegraphics[height=6.5cm]{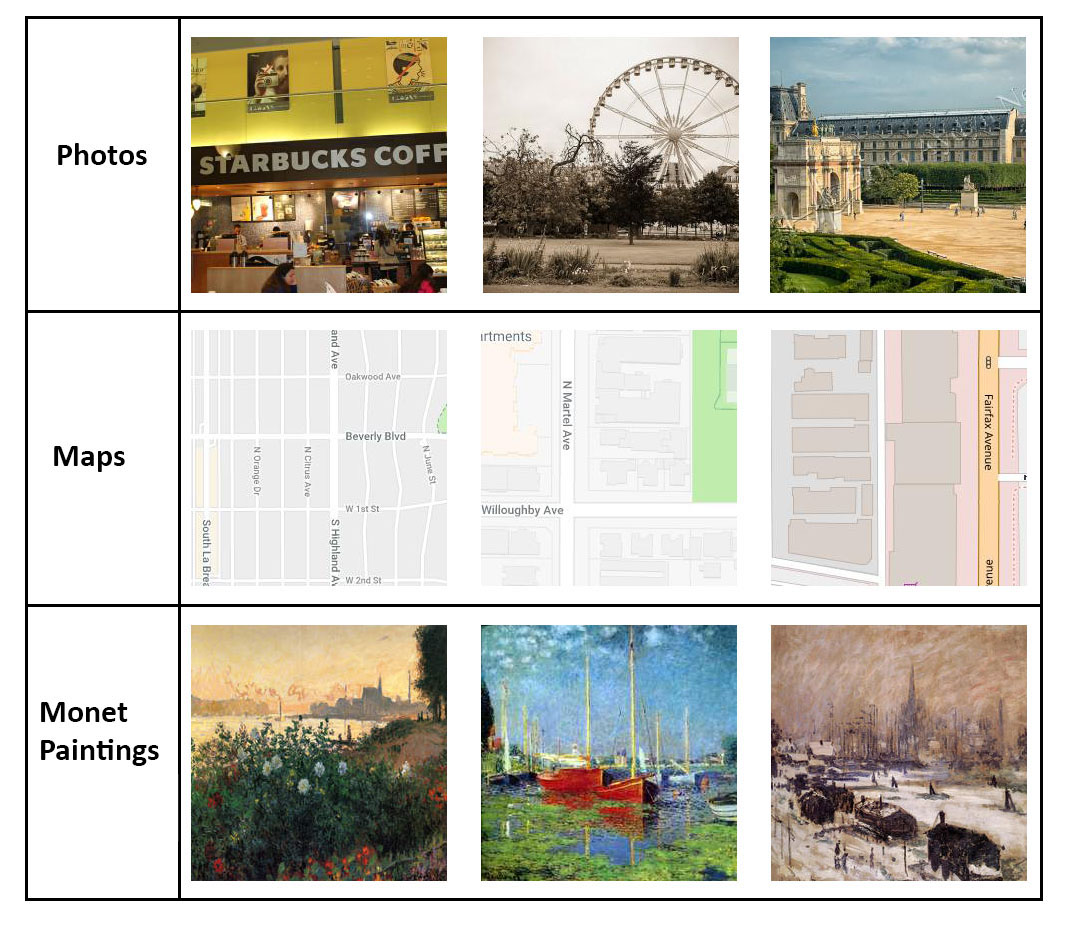}
	\caption{Categories of dataset for training and testing. The first row shows examples of photos collected from \textit{Flickr}, the second row shows examples of tiled images from both \textit{Google Maps} and \textit{OpenStreetMap}, the last line show examples of Monet paintings.}
	\label{fig:dataset}
\end{figure}

We created two categories for preparing the training dataset for the \textit{IsMap} classifier: maps and photos. We randomly selected those styled map tiles collected in Section \ref{sec:framework} as positive samples. 
We did not include the map tiles used for training the classifier in the style transferring process. In addition, we randomly collected \textit{Flickr} photos from its search API\footnote{\url{https://www.flickr.com/services/api/flickr.photos.search.html}} without map content as negative samples. We resized all maps and photos into consistent 256*256 pixel images (Figure \ref{fig:dataset}).

%We created two categories for preparing the training dataset for the IsMap classifier: maps and photos. We randomly selected styled map tiles collected in Section \ref{sec:framework} as positive samples.
%We did not include the map tiles used for training the classifier in the style transferring process. In addition, we randomly collected Flickr photos from its search API\footnote{\url{https://www.flickr.com/services/api/flickr.photos.search.html}} without map content as negative map samples (Figure \ref{fig:dataset}.
%Moreover, to reduce the bias of the model training process with in-balanced number of map images, the ratio between positives and negatives is about 1:1.
%All maps and photos are resized into 256*256 pixels.

\subsection{Evaluation}
After training the aforementioned two C-GAN models, we evaluated the performance of each model based on the \textit{IsMap} classifier. The \textit{IsMap} classifier returns four results:  true positive (\textit{TP}), true negative (\textit{TN}), false positive (\textit{FP}), and false negative (\textit{FN}).
\textit{TP} indicates the number of \textit{transfer styled maps} correctly classified as a map and \textit{TN} indicates the number of tested photos correctly classified as a photo. \textit{FP} indicates the number of testing photos incorrectly predicted as maps and \textit{FN} indicates the number of \textit{transfer styled maps} incorrectly predicted as photos. We then calculated the following four metrics based on the \textit{IsMap} output:
\begin{enumerate}
      \item \textit{Precision}: The portion of the transfer styled images correctly labeled as maps in all output maps, using the following equation:
      \begin{equation}
           Precision = \frac{TP}{TP+FP} 
      \end{equation}
      \item \textit{Recall}: The portion of true positives captured in classification compared to all actual maps in the labeling process, using the following equation:
      \begin{equation}
           Recall = \frac{TP}{TP+FN} 
      \end{equation}
      \item \textit{Accuracy}: The portion of images labeled correctly either as maps or as photos, using the following equation:
      \begin{equation}
           Accuracy = \frac{TP+TN}{TP+TN+FP+FN} 
      \end{equation}
      \item \textit{F1 score}: Combines both precision and recall values to measure the overall accuracy, using the following equation:
      \begin{equation}
           F1 = 2*\frac{Precision*Recall}{Precision+Recall} 
      \end{equation}
\end{enumerate}

\section{Experiment and Results} \label{sec:experiment}
\subsection{Input Datasets}
%\textcolor{red}{
We conducted experiments using the \textit{OSM} raw vector data as well as \textit{Google Maps} with two C-GAN models for two U.S. metropolitan areas to test the feasibility and accuracy of our framework. %}
We utilized the same training and testing datasets for both the \textit{Pix2Pix} and \textit{CycleGAN} models to compare their performance. Because \textit{OSM} vector data coverage varies considerably across regions, we focused on two major cities with high quality data: Los Angeles and San Francisco. We downloaded the \textit{OSM} vector data for these cities from Geofabrik\footnote{\url{https://www.geofabrik.de/data/download.html}} and served the \textit{simple styled maps} as map tiles using \textit{TileCache} and \textit{GeoServer}. To simplify the experiment further, we generated map tiles at only two zoom levels 15 and 18, matching the spatial resolution with the \textit{target styled maps} from \textit{Google Maps}. In total, we generated 870 \textit{simple styled maps} tiles at zoom level 15, and 9,156 image tiles at zoom level 18 for use as the C-GAN training sets. We paired the simple-style maps with the equivalent \textit{Google Maps} tiles for the \textit{Pix2Pix} model.

%\textcolor{red}{
After finishing the training process using the two C-GAN models, we randomly selected 217 and 257 \textit{simple styled maps} tiles, from zoom levels 15 and 18 respectively, to receive the transferred style as testing cases. These selected \textit{simple styled maps} tiles were not included in the C-GAN training process, and thus did not influence style learning and only used for validation.%} 
To train the \textit{GoogleNet}-based CNN classifier \textit{IsMap}, we downloaded 5,500 photos without map content from \textit{Flickr}, and 500 tiled  maps from both \textit{Google Maps} and \textit{OSM} styled maps at different zoom levels. We then trained the \textit{IsMap} classifier using this sample to produce the binary label of True or False.

\subsection{\textit{Pix2Pix}: Style Transferring with Paired Data}
\begin{figure*}
	\centering
	\includegraphics[height=6cm]{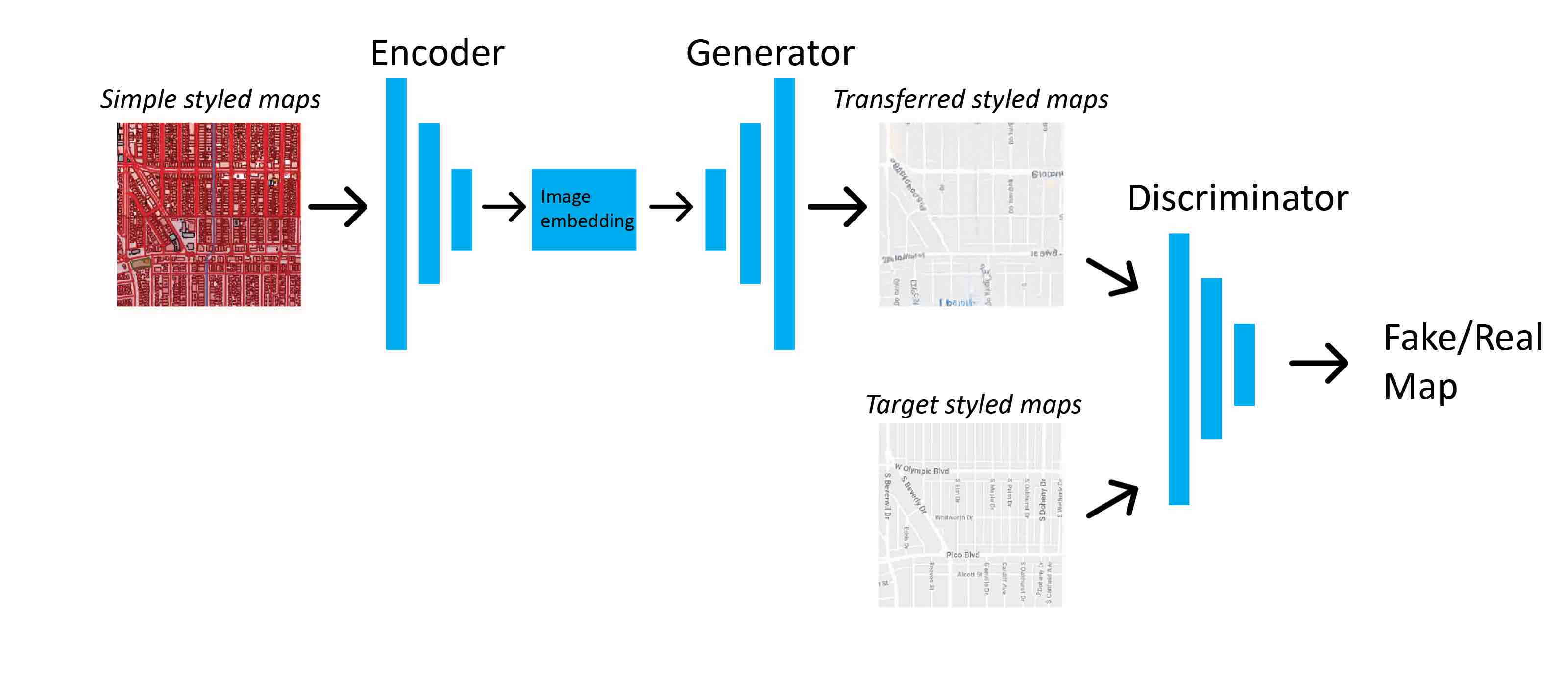}
	\caption{Data flow of \textit{Pix2Pix} in this research.}
	\label{fig:pix2pixworkflow}
\end{figure*}

\begin{figure}
	\centering
	\includegraphics[height=8.5cm]{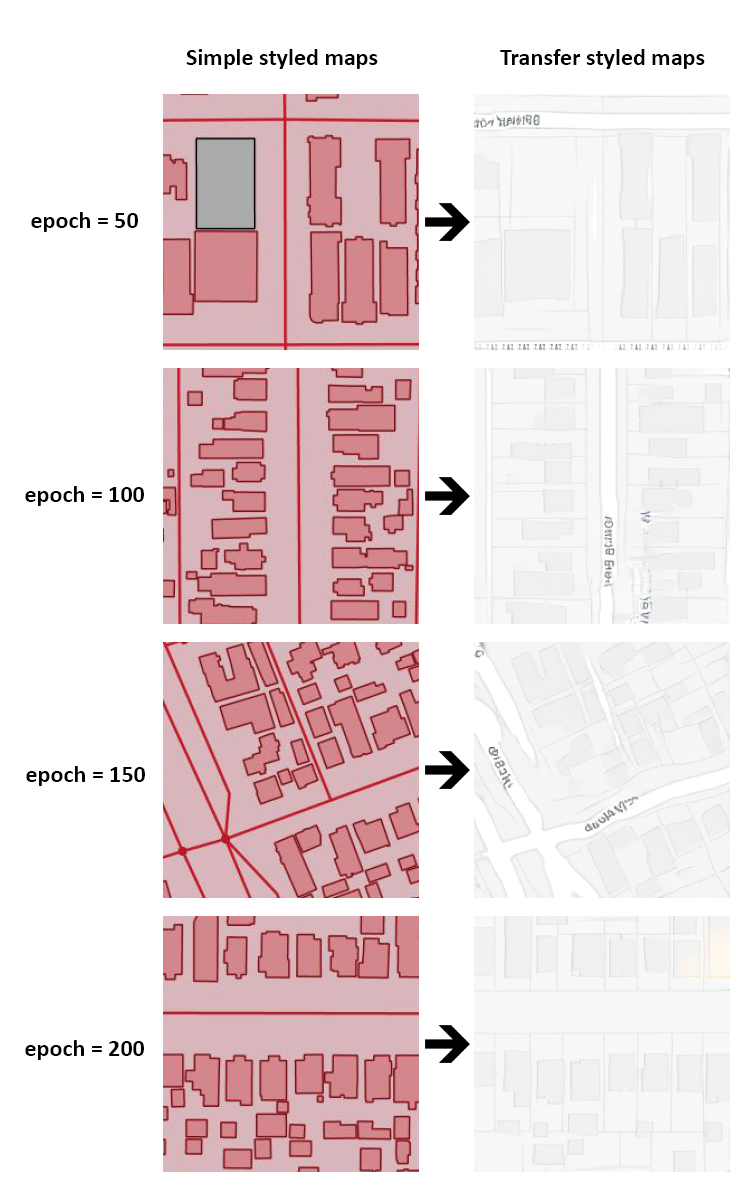}
	\caption{Training process using \textit{Pix2Pix} at zoom level 18. Examples of \textit{simple styled maps} and \textit{target styled maps} in epoch 50, 100, 150 and 200 are shown individually.}
	\label{fig:pix2pix18epoch}
\end{figure}

\begin{figure}
	\centering
    \includegraphics[height=8.5cm]{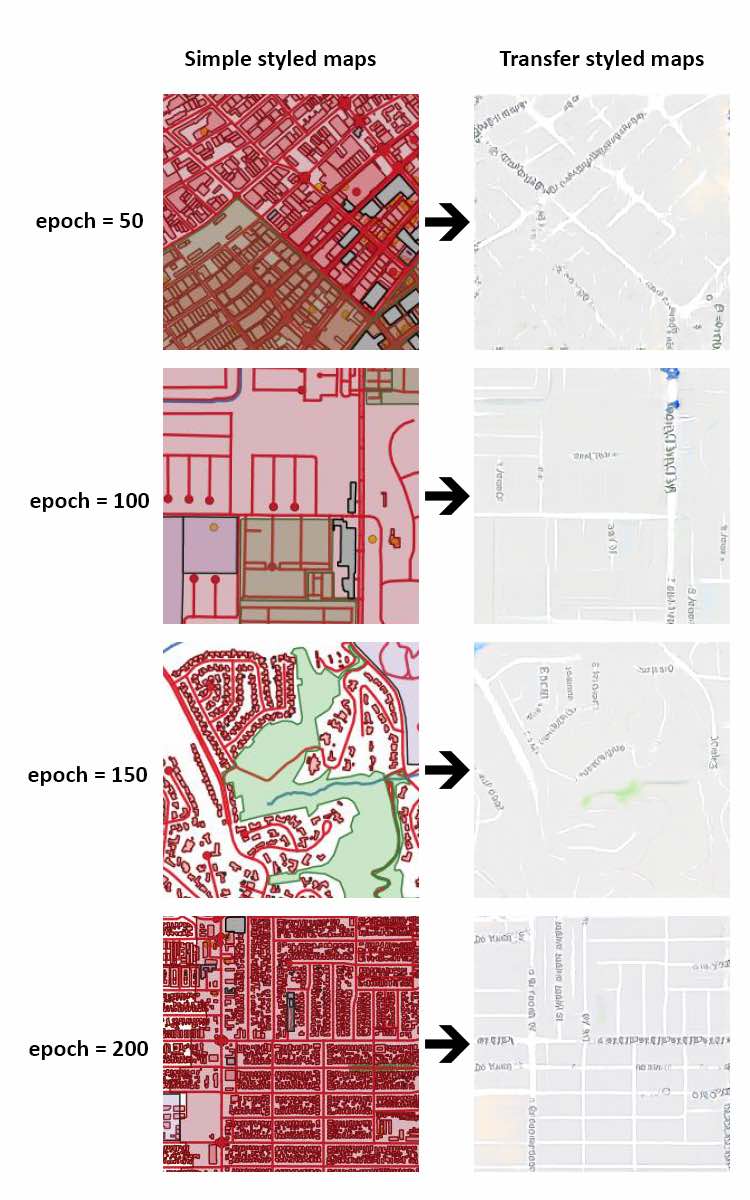}
    \caption{Training process using \textit{Pix2Pix} at zoom level 15. Examples of \textit{simple styled maps} and \textit{target styled maps} in epoch 50, 100, 150 and 200 are shown individually.}
    \label{fig:pix2pix15epoch}
\end{figure}

\begin{figure}
	\centering
	\includegraphics[height=6cm]{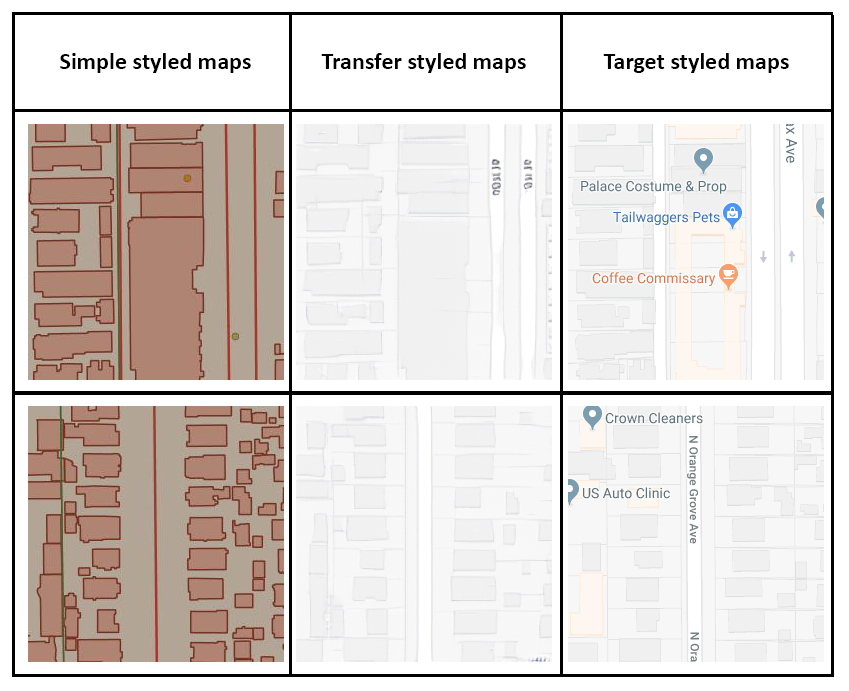}
	\caption{Results of the map style Transfer using \textit{Pix2Pix} at zoom level 18. Examples of \textit{simple styled maps} are shown in the first column, \textit{transfer styled maps} based on \textit{Pix2Pix} are shown in the second column, and the \textit{target styled maps} from \textit{Google Maps} are shown in the last column.}
	\label{fig:pix2pix18}
\end{figure}

\begin{figure}
	\centering
	\includegraphics[height=6cm]{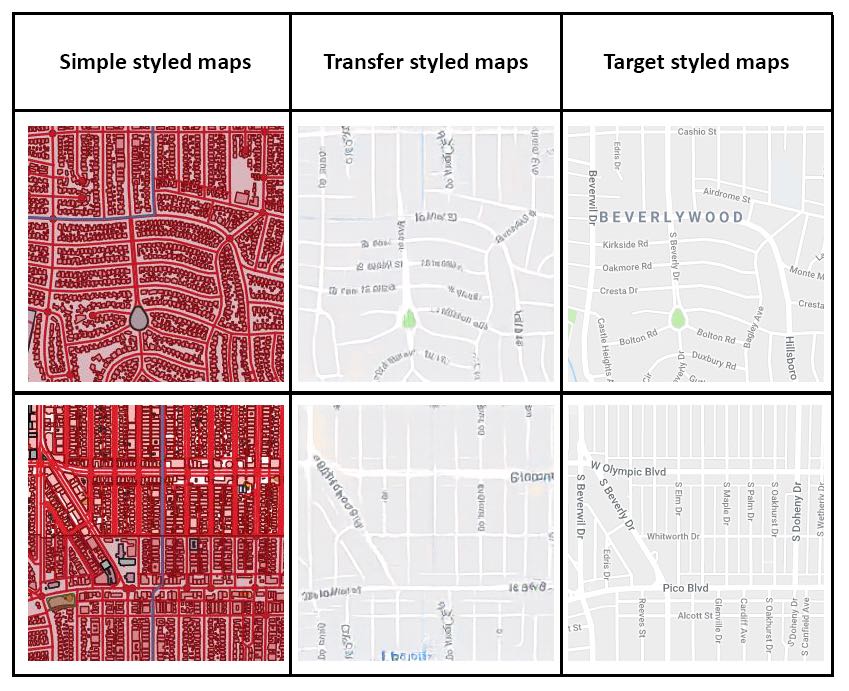}
	\caption{Results of the map style transfer using \textit{Pix2Pix} at zoom level 15. Examples of \textit{simple styled maps} are shown in the first column, \textit{transfer styled maps} based on \textit{Pix2Pix} are shown in the second column, and the \textit{target styled maps} from \textit{Google Maps} are shown in the last column.}
	\label{fig:pix2pix15}
\end{figure}

%\textcolor{red}{
	Figure \ref{fig:pix2pixworkflow} illustrates the training process for the \textit{Pix2Pix} model. First, we created the \textit{simple styled maps} tiles from the \textit{OSM} vector data, and then fed these tiles into the generator \(G\) by encoding and embedding those images as vectors to generate the ``fake'' \textit{transfer styled maps}. Then, we input the generated \textit{target styled maps} and the \textit{transfer styled maps} into the discriminator \(D\), which iterated through 200 epochs until the discriminator no longer delineate the real versus fake maps. Figures \ref{fig:pix2pix18epoch} and \ref{fig:pix2pix15epoch} provide examples of \textit{Pix2Pix} \textit{transfer styled maps} tiles at 50, 100, 150, and 200 epochs for zoom level 18 and 15 respectively.
%}

\subsubsection{Generative Process with Map Tiles at a Large Scale}
%\textcolor{red}{
Figure \ref{fig:pix2pix18} depicts \textit{Pix2Pix} \textit{transfer styled maps} generated at zoom level 18. Intuitively, the \textit{transfer styled maps} look similar to the \textit{Google Maps} tiles, which proves the basic feasibility of our AI framework broadly and the utility of \textit{Pix2Pix} C-GAN model specifically. Compared with the original \textit{simple styled tiled maps}, \textit{Pix2Pix} preserves the detailed geometry of the roads and buildings with minimal observed generalization, but fails to maintain legible labels and colored markers (see discussion below). Notably, \textit{Pix2Pix} consistently applies the target white road styling with a consistently line thickness to input line features and also applies the target grey building styling with rigid corners to input rectangular features, showing a relationship between salient visual variables and feature types in the transfer style generative process.
%}
%Figure \ref{fig:pix2pix18} illustrates the transferred-styled maps generated at zoom-level 18.
%Intuitively, those transferred-styled maps look similar to the Google Maps tiles, which proves the feasibility of the model. More specifically, roads are generated from line features and buildings are generated from the rectangles.
%35242344Compared with the original simple-styled tiled maps, the geometry of the roads and buildings are reserved well with much detailed information.

\subsubsection{Generative Process with Tiles at a Small Scale}
%\textcolor{red}{
	Figure \ref{fig:pix2pix15} shows \textit{Pix2Pix} \textit{transfer styled maps} at zoom level 15. Compared with the original \textit{simple styled maps}, \textit{Pix2Pix} overgeneralizes the geometry of road features at zoom level 15, with many intermediate-size streets removed from the \textit{target styled maps}. This overgeneralization potentially is explained by differences in the vector data schemas between \textit{OSM} and \textit{Google Maps}, resulting in a thinner road network in the \textit{OSM}-based \textit{transfer styled maps}. In comparison, \textit{Pix2Pix} appropriately thinned the building features in the \textit{transfer styled maps}, eliminating most buildings at zoom level 15 compared to zoom level 18 based on differences in the \textit{target styled maps}. Thus, \textit{Pix2Pix} is reasonably successful at multiscale generalization and style transfer, with the \textit{Pix2Pix} model preserving feature types from the \textit{simple styled maps} that are most salient in the \textit{target styled maps} when changing zoom levels. 
%}
%Figure \ref{fig:pix2pix15} shows the results of transferred-styled maps at zoom-level 15.
%Compared with the original simple-styled tiled maps, the geometry of geographic features can be learned and generated by the model as well.
%However, different from the results at level 18 where all buildings are preserved, buildings are generalized at level 15 as it is at a smaller map scale.
%Although roads maintain their geometry, most buildings are eliminated except for those which occupied large spaces can be preserved.

\subsubsection{Limitations of the \textit{Pix2Pix}}
%\textcolor{red}{
Although the generated \textit{transfer styled maps} are similar to the \textit{target styled maps} at both zoom levels, several limitations in the output exist. First, the map labels are important parts of the \textit{target styled maps}, but are not preserved in the \textit{transfer styled maps}, a major shortcoming of the image-based \textit{Pix2Pix} model. Thus, while labeling and annotation falls outside of style learning, \textit{Pix2Pix} does still learn knowledge of where to put the labels on the map for subsequent manual label placement. Second, \textit{Pix2Pix} does not preserve less frequently observed colors in the \textit{target styled maps}, such as the colored markers in Figure \ref{fig:pix2pix18}, instead basing the style transfer on the most common colors in the target style. Accordingly, the \textit{transfer styled maps} does not capture grassland, lakes, etc., compared to the road and building colors dominating the urban landscapes of Los Angeles and San Francisco. However, color sensitivity may improve when expanding the geographic extent, and thus feature diversity, of the tilesets. 
%}

%Although the generated maps seem to be pretty similar to the target-styled maps, several limitations exist. First, the map labels and markers are important parts of the target-styled maps. However, it caused problems in the training process of this research.
%From the Figure \ref{fig:pix2pix18} and Figure \ref{fig:pix2pix15}, it can be referred that without auxiliary text information, the generated labels are unintelligent texts without valuable information. But the knowledge of where to put the labels on the map can still be learned from the target-styled maps. Another issue is about the colorization.
%Ideally, based on the input color scheme for different types of geographic features, the model should be able to distinguish them. However, the outputs are not so sensitive to the color.  For regions of grassland, lakes, etc., only a few are rendered in matched colors correctly.

\begin{figure*}
	\centering
	\includegraphics[height=7.5cm]{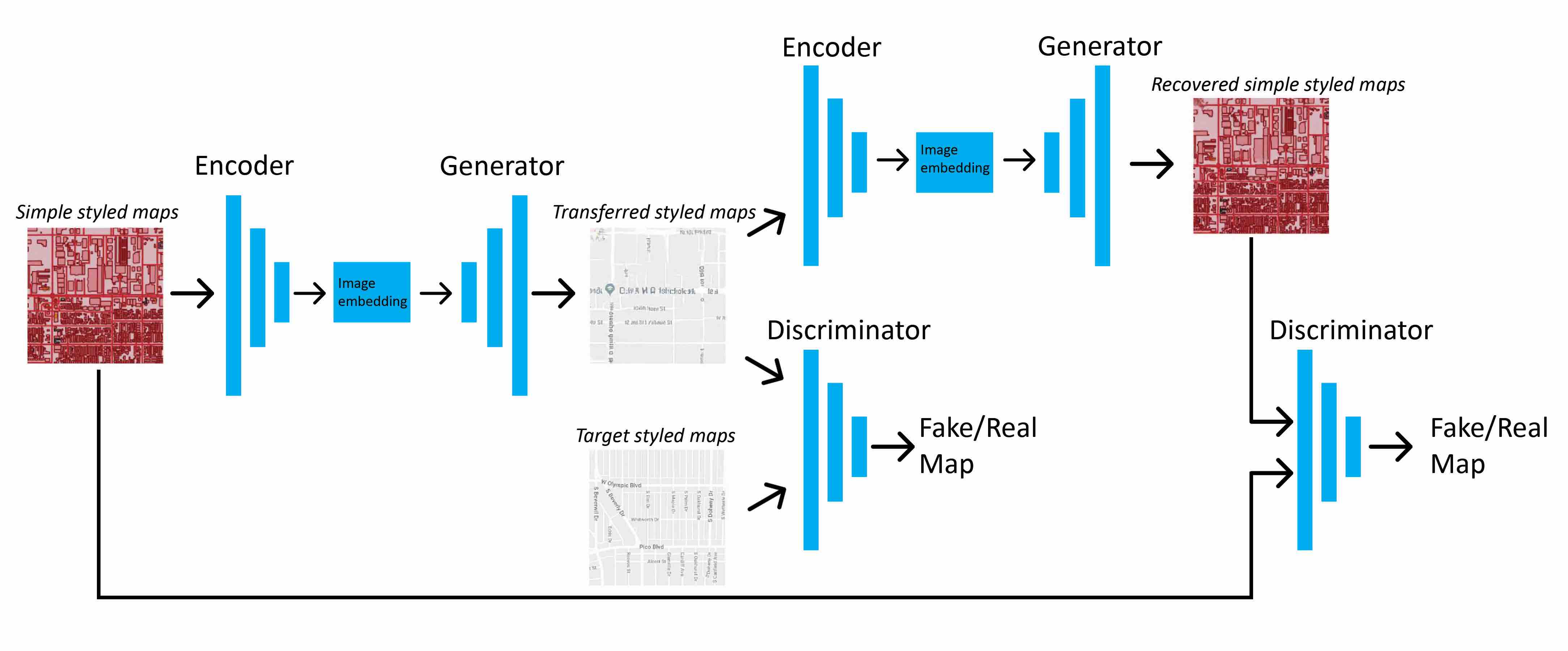}
	\caption{Data flow of \textit{CycleGAN} in this research.}
	\label{fig:cycleganworkflow}
\end{figure*}

\begin{figure}[ht]
	\centering
	\includegraphics[height=8.5cm]{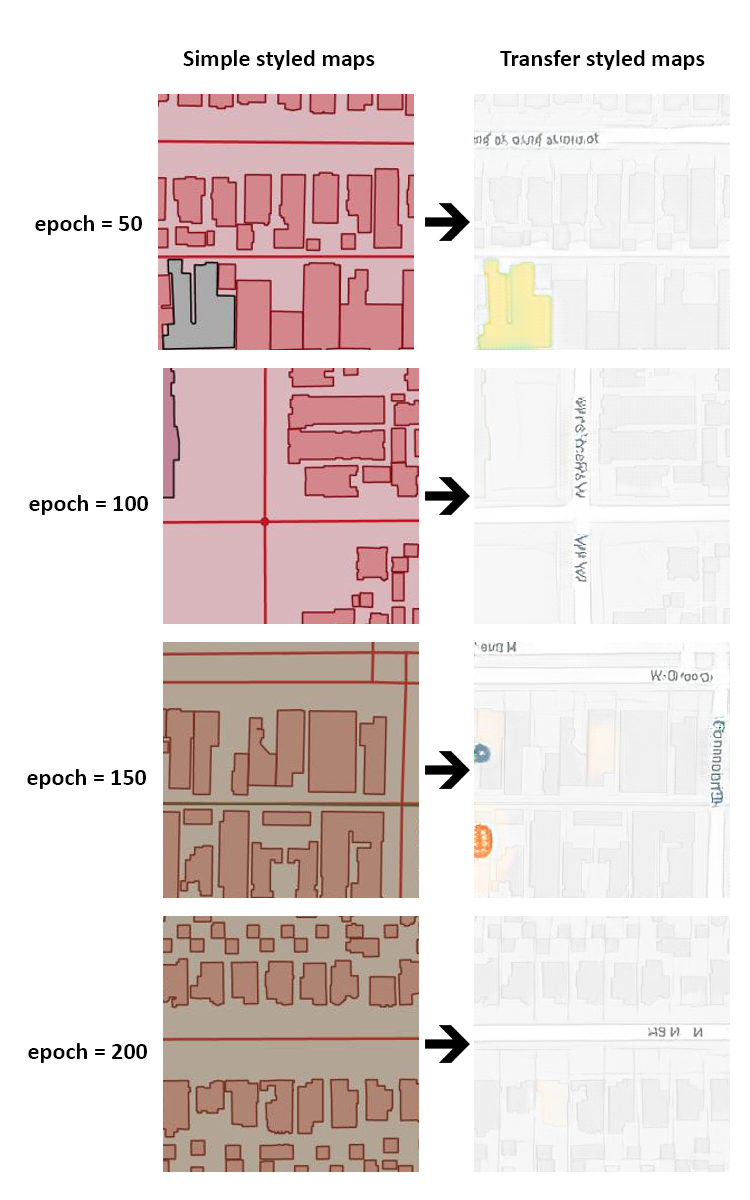}
	\caption{Training process using \textit{CycleGAN} at zoom level 18. Examples of \textit{simple styled maps} and \textit{target styled maps} in epoch 50, 100, 150 and 200 are shown individually.}
	\label{fig:cyclegan18epoch}
\end{figure}

\begin{figure}[ht]
	\centering
	\includegraphics[height=8.5cm]{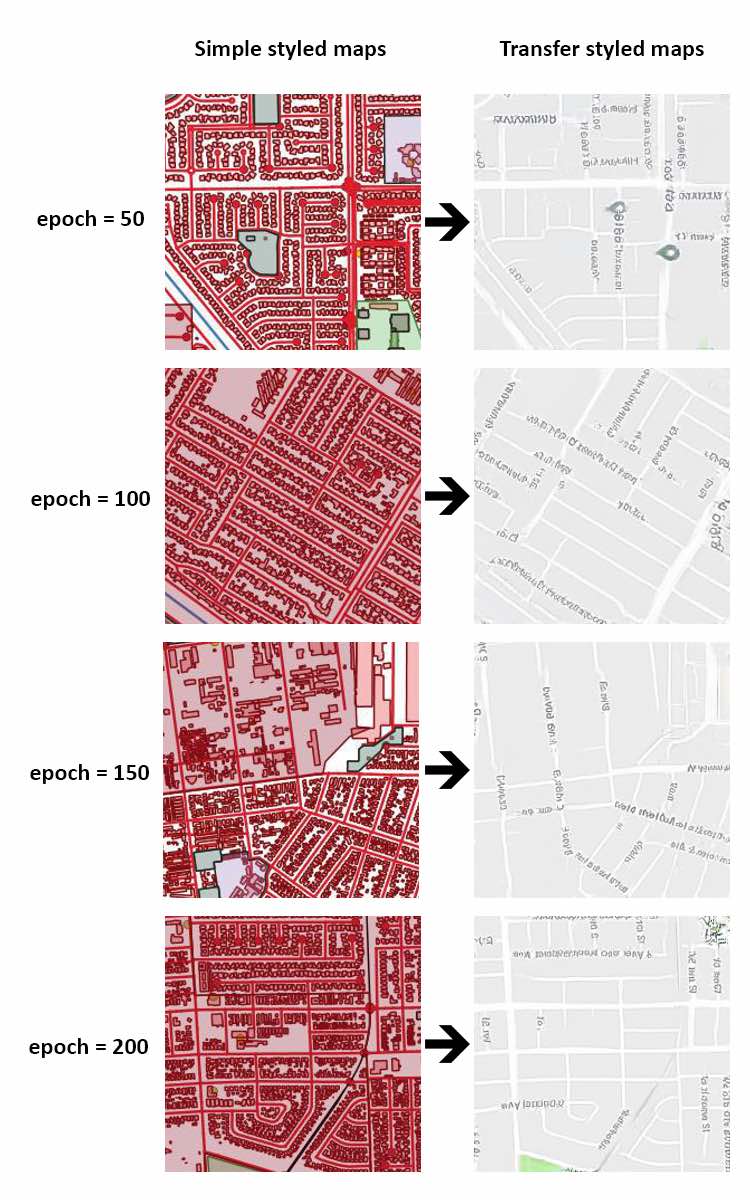}
	\caption{Training process using \textit{CycleGAN} at zoom level 15. Examples of \textit{simple styled maps} and \textit{target styled maps} in epoch 50, 100, 150 and 200 are shown individually.}
	\label{fig:cyclegan15epoch}
\end{figure}

\begin{figure}[ht]
	\centering
	\includegraphics[height=6.5cm]{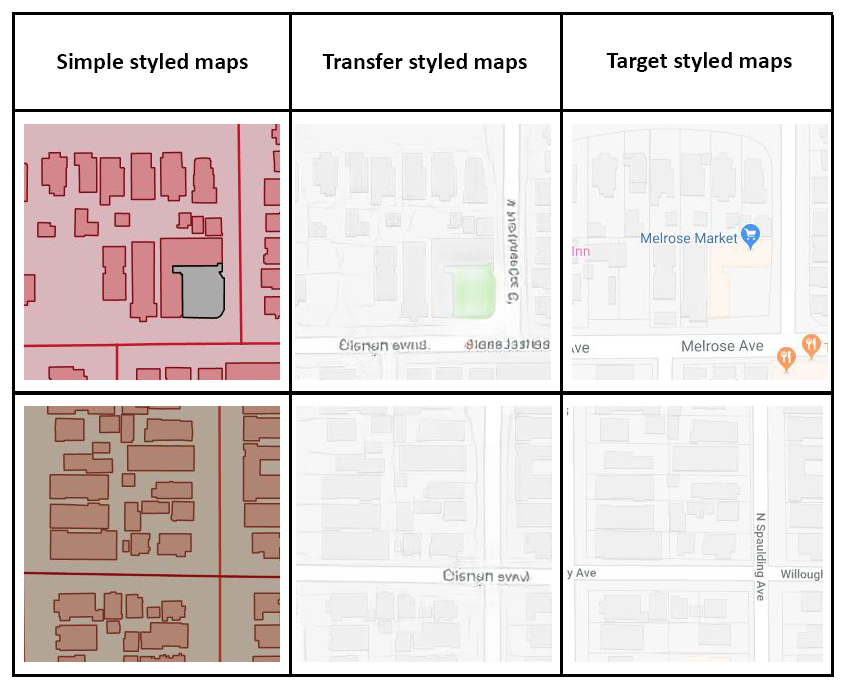}
	\caption{Results of the map style transfer using \textit{CycleGAN} at zoom level 18. Examples of \textit{simple styled maps} are shown in the first column, \textit{transfer styled maps} based on \textit{CycleGAN} are shown in the second column, and the \textit{target styled maps} from \textit{Google Maps} are shown in the last column.}
	\label{fig:cyclegan18}
\end{figure}

\begin{figure}[ht]
	\centering
	\includegraphics[height=6.5cm]{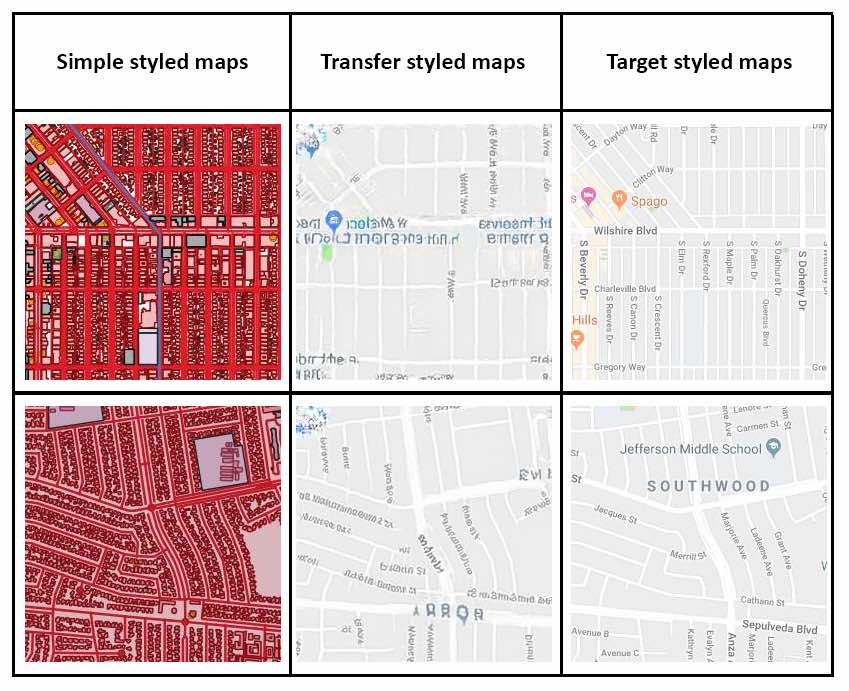}
	\caption{Results of the map style transfer using \textit{CycleGAN} at zoom level 15. Examples of \textit{simple styled maps} are shown in the first column, \textit{transfer styled maps} based on \textit{CycleGAN} are shown in the second column, and the \textit{target styled maps} from \textit{Google Maps} are shown in the last column.}
	\label{fig:cyclegan15}
\end{figure}

\subsection{\textit{CycleGAN} Style Rendering with Unpaired Data}
%\textcolor{red}{
Figure \ref{fig:cycleganworkflow} illustrates the training process for the \textit{CycleGAN} model. Similar to the \textit{Pix2Pix} training, we encoded the \textit{OSM} vector data to create the \textit{simple styled maps} tiles. These \textit{simple styled maps} are used as input to generator \(G\) to produce \textit{transfer styled maps} and also represent knowledge that can be restored for discriminator \(D\). Again, \textit{CycleGAN} does not used paired data, with the \textit{target styled maps} generated using randomly selected images from \textit{Google Maps}. Similar to the \textit{Pix2Pix} model training, we also trained the \textit{CycleGAN} model across 200 epochs, enabling comparison performance between the two models. Figures \ref{fig:cyclegan18epoch} and \ref{fig:cyclegan15epoch} provide examples of \textit{CycleGAN} \textit{transfer styled maps} tiles at 50, 100, 150, and 200 epochs for zoom level 18 and 15 respectively.
%}

%\textcolor{red}{
	%The training process of CycleGAN is illustrated in Figure \ref{fig:cycleganworkflow}.
	%Similar to the \textit{Pix2Pix}, simple styled maps will be encoded as vectors which contain knowledge to generate transferred styled maps.
	%Those generated maps will be input into the other similar structure as input knowledge to recover the original simple styled maps.
	%By generating fake maps so that two \(D\) can not identify real and fake images, the style of the target styled maps can be learned.
	%Please note that the target styled maps in \textit{Pix2pix} are paired data with simple styled maps while are different and randomly selected in CycleGAN.
%}

%\textcolor{red}{
%Similar to the Pix2Pix model training in which we take 200 epochs, for CycleGAN model, we also trained data in 200 epochs,
%there is no specific epoch limit in our selected implementation of the CycleGAN model. 
%For each training epoch in Pix2Pix, all input images as well as the paired images are passed to generate transferred-styled maps through the training process.
%Whereas in CycleGAN, only one simple-styled image and one target-styled map image are randomly selected and used in one step for the training. 
%so that we can 
%trained the model for about 120,000 steps to 
%compare the performance at the same zoom levels (Figure\ref{fig:cyclegan18epoch} and \ref{fig:cyclegan15epoch}).
%}

\subsubsection{Generative Process with Tiles at a Large Scale}
%\textcolor{red}{
Figure \ref{fig:cyclegan18} shows \textit{CycleGAN} \textit{transfer styled maps} generated at zoom level 18. \textit{CycleGAN} preserves the shapes of roads and building well. Notably, \textit{CycleGAN} includes marker overlays from the target styled maps in some of the transfer styled maps, a benefit over \textit{Pix2Pix}, although the colors and locations are incorrect. Like \textit{Pix2Pix}, \textit{CycleGAN} failed to apply legible text from the target styled maps. Finally, a broader range of colors are included in the \textit{CycleGAN} transfer styled maps compared to the \textit{Pix2Pix} output, although the coloring is not applied to the correct locations (e.g., the highlighted Melrose Market in Figure \ref{fig:cyclegan18}).
%}

%\textcolor{red}{
%Figure \ref{fig:cyclegan18} shows the transferred-styled maps generated at zoom-level 18.
%It can be shown that the geometric features are remained in the generated tiled maps. 
%Both roads and building preserve their shapes well.
%\textcolor{red}{
%Compared with the results based on Pix2Pix, markers can be generated although their locations might be wrong.
%And more annotation texts can be generated although most texts are still intelligent.
%}

\subsubsection{Generative Process with Tiles at a Small Scale}
%\textcolor{red}{
Figure \ref{fig:cyclegan15} depicts CycleGAN \textit{transfer styled maps} generated at zoom level 15. Results show that the skeletons of the roads are remained. Similar to the results in \textit{Pix2Pix} at zoom level 15, buildings are generalized at this level. Markers again are generated, with the marker shape relatively well preserved. Many different features types are distinguishable in the output results, including primary and secondary roads, building footprints, and less common features such as grasslands. Most generated maps look in realistic.
%}

%Figure \ref{fig:cyclegan15} shows the transferred-styled maps generated at zoom-level 15.
%Results show that the skeletons of the roads are remained.
%\textcolor{red}{
%Similar to the results in Pix2Pix at zoom-level 15, buildings are also generalized at this level. 
%And the markers can be generated with shapes well preserved.
%Different types of map features like highway roads and general roads, grasslands and building lands can be distinguished in the output result.
%Maps generated look in realistic.
%}
%Those generated maps are not as clear as the results in the training process.

\subsubsection{Limitations of the CycleGAN}
Although \textit{CycleGAN} can generate maps with with a similar style to the \textit{target styled maps}, challenges remained. Similar to the results from \textit{Pix2Pix}, the generated labels are illegible and do not contain valuable information. Although \textit{CycleGAN} does generate marker overlays in the appropriate shape, the color and location of the markers are incorrect. Compared to \textit{Pix2Pix}, \textit{CycleGAN} inconsistently applies line widths (sizes) to features like roads and the directions (orientations) of roads change considerably from the \textit{simple styled maps}, a major hindrance to the usability of the resulting maps.

%\textcolor{red}{
%Though CycleGAN can generate maps with similar style of the target-styled maps, challenges remained.
%Similar to the results from Pix2Pix, the generated labels are unintelligent texts and do not include valuable information.
%Though the markers can be generated, the color of the markers are incorrect. 
%The width of the line features like roads are not stable, that means the width may vary along with the roads.
%And the direction of the roads might have slightly change compared with the simple-styled maps.
%}
%Also, the color rendering of some regions might not accurate as the rendered color
%In sum, CycleGAN can only preserve the geometry skeletons while other attributes, like labels, markers are removed in our experiments.  In addition, the results of CycleGAN are not effective in capturing feature color settings either. The dominant hue of each map is almost the same.
%And different types of spatial features cannot be distinguished. It might be resulted from a different experiment setting compared with the original implementation in \cite{CycleGAN2017}. 

\subsection{Evaluation and Comparison}
\begin{figure}
	\centering
	\includegraphics[height=5.5cm]{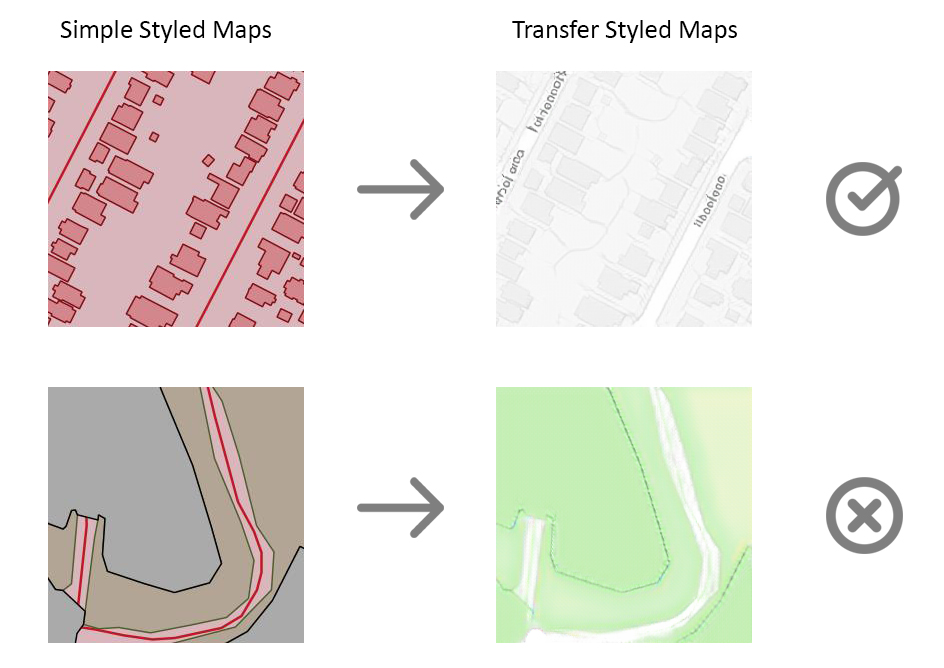}
	\caption{Examples of binary classification results of \textit{IsMap} classifier. The \textit{transfer styled maps} in above figure is classified as map correctly while in bottom is wrongly classified as photo.}
	\label{fig:predictionresult}
\end{figure}

%\textcolor{red}{
Qualitatively, the \textit{transfer styled maps} by \textit{CycleGAN} are visually similar to those generated by \textit{Pix2Pix}, with several notable differences by visual variable, feature type, and zoom level. Table \ref{tab:tab1} presents comparative quantitative results using measures derived from the \textit{IsMap} classifier, including the aforementioned precision, recall, accuracy, and F1-score. For reference, Figure \ref{fig:predictionresult} provides two outcomes of the \textit{IsMap} classifier from the experiment: a \textit{transfer styled maps} tile classified as a map and one rejected as a map. As shown in Table \ref{tab:tab1}, the \textit{CycleGAN} performs better than the \textit{Pix2Pix} in transferring the map styles from \textit{target styled maps} to \textit{simple styled maps}. The higher the evaluation metrics is, the better the model is for map style transfer. The F1-scores of \textit{CycleGAN} in both zoom levels 15 and 18 are higher than that of \textit{Pix2Pix}. The \textit{transfer styled maps} generated at zoom level 15 are more realistic with the F1-score 0.998 compared with results at zoom level 18 with F1-score 0.841 using \textit{Pix2Pix}. The result is similar for \textit{CycleGAN}, in which the quality of \textit{transfer styled maps} at zoom level 15 with F1-score 1.0 is better than that at zoom level 18 with F1-score 0.95. Hence, it can be concluded that zoom level 15 is more suitable for generating \textit{transfer styled maps} in this study, an important finding pointing to the feasibility of AI broadly and GANs specifically to assist with multiscale generalization and styling.
The results also demonstrate that the \textit{CycleGAN} model is more effective for the map style transfer task at both zoom levels 15 and 18 compared with the \textit{Pix2Pix} model.

\begin{table}[h]
    \centering
    \begin{tabular}{|c|c|c|c|c|}
    \hline
    Model     & \multicolumn{2}{c|}{Pix2Pix} & \multicolumn{2}{c|}{CycleGAN} \\ \hline
    Data      & Level 15      & Level 18     & Level 15      & Level 18      \\ \hline
    Precision & 1.000              &0.989              &1.000               &0.992               \\ \hline
    Recall    & 0.995              &    0.732          &    1.000           &     0.911          \\ \hline
    Accuracy    &   0.998            &      0.862        &      1.000        &       0.951        \\ \hline
    F1-Score    &       0.998        &          0.841    &          1.000     &           0.950    \\ \hline
    \end{tabular}
    \caption{Evaluation metrics of two GAN models at different map zoom levels}
\label{tab:tab1}
\end{table}

\section{Discussion} \label{sec:discussion}

\begin{figure}
    \centering
        \includegraphics[height=6cm]{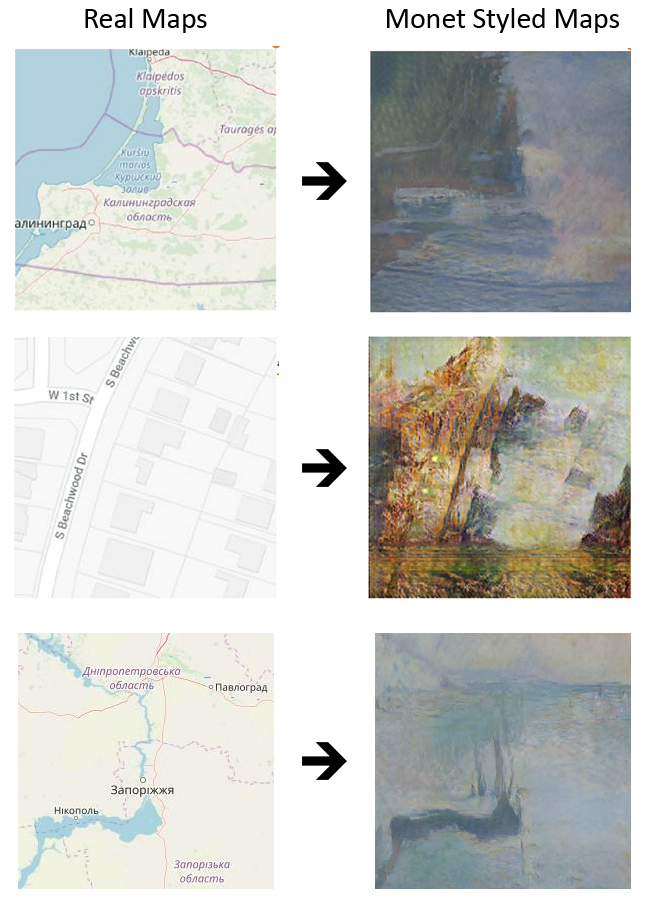}
        \caption{Results with transferred Monet painting styles using \textit{CycleGAN}. The left column is real \textit{OSM} maps, and the right column is the maps with Monet painting styles transferred. Results are not good. }
        \label{fig:styletransfer}
\end{figure}
The results of our experiments with the Google Maps style are encouraging, and generate several insights into future research at the intersections of AI and cartographic design. First, we explored if non-map input also might work for map style transfer using GANs. As an example, we downloaded an artwork library by Claude Monet—an impressionist painter with an aesthetic style characterized by vivid use of color and dramatic interplay of light and shadow—for use as a target painting style. Figure \ref{fig:dataset} shows several Monet examples used as the target painting style. While Monet primarily painted landscapes, there is no georeferenced information in the downloaded Monet artwork library and thus requires the unpaired CycleGAN model for style transfer. We again employed OSM for the simple styled maps receiving the target painting style. Figure \ref{fig:styletransfer} shows several transfer styled maps generated by CycleGAN using the Monet target style. Qualitatively, the output transfer styled maps appear to resemble paintings more than maps, although some map-like shapes and structures emerge. To confirm our visual interpretation, we again imported the Monet inspired transfer styled maps into the IsMap classifier and then used a modified deep-CNN classifier to categorize the images as photos, maps, or (new to the modified classifier) paintings. Less than 1$\%$ of the transfer styled maps are classified as maps, with most instead classified as paintings. One possible reason for the poorer results is less intensive training on the limited set of input visual art compared to the voluminous map tilesets. Therefore, transferring styles from visual art to maps might not be effective using the current workflow and requires further research.

\begin{figure}
	\centering
	\includegraphics[height=10cm]{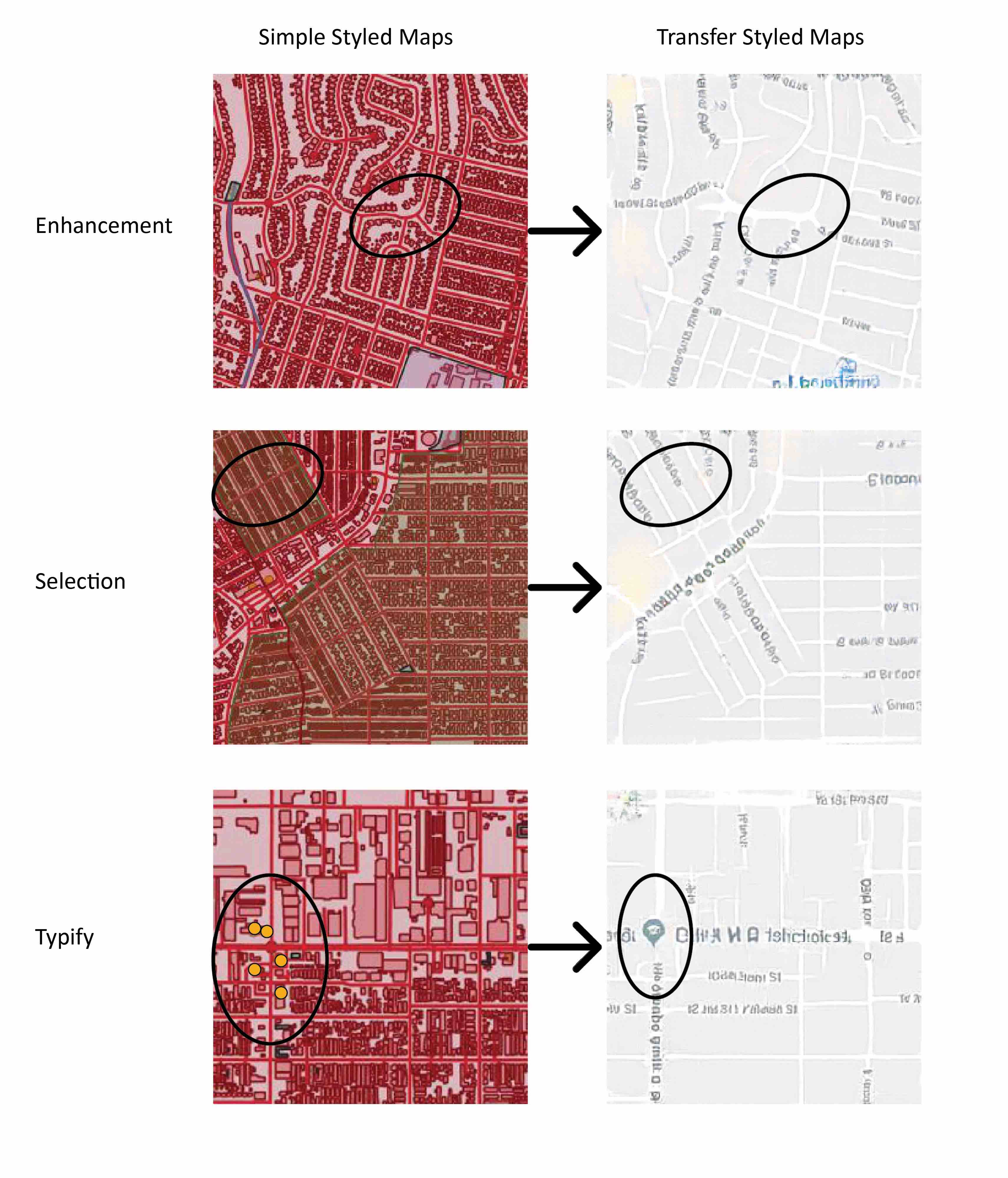}
	\caption{Examples of map generalization. The first row shows enhancement as the road in the black circle thickened; the second row shows selection as the road in the black circle is not selected in the \textit{transfer styled maps}; the last row shows typify as several POIs are represented as one marker.}
	\label{fig:mapgeneralization}
\end{figure}

Second, we took a deeper look at the way that the GANs generalize linework from the input simple styled maps in the resulting transfer style maps. Figure \ref{fig:mapgeneralization} shows how the GAN transfer effectively approximates several generalization operators (including enhancement, selection, and typify) to mimic the target Google Maps style at zoom level 15. First, enhancement of the road width is applied to create an artificial road hierarchy present across the target styled maps but not included in the original simple styled maps (Figure \ref{fig:mapgeneralization}, top). Based on our analysis, the distance from buildings to roads is the primary spatial structure that affects how the GANs apply the enhanced road weight. Second, many roads are selectively eliminated from the simple styled maps in the resulting zoom level 15 transfer styled maps (Figure \ref{fig:mapgeneralization}, middle). Roads that do not follow the orientation of the general street network are more likely candidates for elimination, such as the circled road running southeast to northwest. Finally, point markers are typified in the transfer styled maps, with the marker placed in the general location of a number of representative points of interest (POIs) from the simple styled maps (Figure \ref{fig:mapgeneralization}, bottom). It is important to note that the GAN models are unlikely applying specific rule-based generalization operators, but rather the resulting transfer styled maps exhibit characteristics of these operators. Studying the generalization operators approximated by GANs may help optimize manual generalization and provide new insights for cartographic design broadly.

\begin{figure}
	\centering
	\includegraphics[height=4cm]{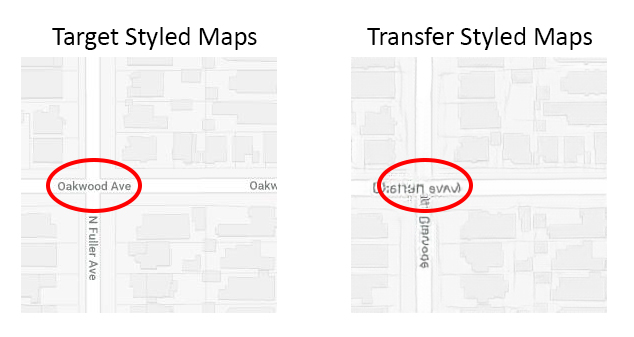}
	\caption{Example of a topology error in this study. Compared with the \textit{target styled maps}, the topology of the road intersection in the \textit{transfer styled maps} is wrong.}
	\label{fig:topology}
\end{figure}

Finally, most existing research about map styling is based on vector data. Vector data records spatial coordinates and feature attributes separately, making it convenient and suitable for geometry generalization and symbol styling. However, our research is based on raster data, as GANs more commonly are used to process images. Each approach has pros and cons. Different map features are stored in different layers of vector data, meaning the styling of each layer is independent from others. As a result, styling of different layers may not work in concert in some places, making it difficult to achieve an optimal set of styles that work cohesively across the ‘map of everywhere’. Additionally, rich vector data requires more computing resources to apply the styles, limiting both design exploration by the cartographer and real-time rendering for the audience. In comparison, raster data—such as the tilesets used in this research—collapse both the spatial and attribute information into a single pixel value. With advanced image processing methods like convolutional neural networks and GANs, it is easier to calculate the output styling for raster tilesets, enabling large volume style transfer without complex style lists. However, the topology relationship of spatial features may break because such raster-based methods only use the single pixel value for computation. For example, Figure \ref{fig:topology} shows one common topology error occurring in our research: road intersections. In the transferred styled maps, the bold roads pass one over another rather than intersect, while in the target styled maps, roads are connected via the intersection. In the future, we plan on exploring techniques to minimize topology errors in the transfer styled maps.

\section{Conclusion and Future Work} \label{sec:conclusion}
In this research, we investigate multiscale map style transfer using state-of-the-art AI techniques. Specifically, two conditional generative adversarial network models, the \textit{Pix2Pix} based on paired data and the \textit{CycleGAN} based on unpaired data, are employed for the cartographic style transfer. The results of two methods show that GANs have the capability to transfer styles from customized styled maps like \textit{Google Maps} to another without \textit{CartoCSS} style sheets.

To answer the three research questions proposed in section 1, the study explored whether the two models can preserve both the complex patterns of spatial features and the aesthetic styles in generated maps. From the qualitative visual analysis, several visual variables of the target styled maps are retained, including the feature color, the line width (size), and feature shape, especially in urban areas with buildings and roads. The locations of some markers and annotations are also learned from the \textit{transfer styled maps}. However, the GANs failed to apply legible text labels from the \textit{target styled maps}. Moreover, we tested the performance of two models at two different zoom levels of the map data with different geographic ranges and feature compositions. In order to check whether the output still appears to be maps, we implemented a deep convolutional neural network to evaluate the results. 

The \textit{CycleGAN} model performs better than the \textit{Pix2Pix} model using quantitative measures in our experiments regardless of the map zoom level. The transfer styled maps at level 15 perform better than that at level 18 using both \textit{Pix2Pix} and \textit{CycleGAN} models. There is a wide gap in performance at zoom level 18, with CycleGAN producing an F1-score of 0.95 but Pix2Pix only reaching a score of 0.841. Thus, geographic features at small scales can be generalized automatically, and important result with positive implications for the use of AI to assist with multiscale generalization and styling. Taken together, these findings prove that GANs have a great potential for map style rendering, transferring, and maybe other tasks in cartography.

Although most generated maps look realistic, several problems and challenges remain. First, since the GAN models are based on pixels from the input images, the topology of geographic features may not be well retained as discussed in Section \ref{sec:discussion}.
Second, because of the existence of point markers and textual labels in the tiled maps, the quality of \textit{transfer styled maps} are influenced by them. However, the markers and text labels are important to the map purpose and might require separate pattern recognition models \citep{chiang2016unlocking} to achieve better results. Therefore, in our future work, we also plan to train maps without labels and markers to reduce the bias caused by them.

In sum, this research demonstrates substantial potential for implementing artificial intelligence techniques in cartography. We outline several important directions for the use of AI in cartography moving forward. First, our use of GANs can be extended to other mapping contexts to help cartographers deconstruct the most salient stylistic elements that constitute the unique look and feel of existing designs, using this information to improve designs in future iterations. This research also can help non-experts who lack professional cartographic knowledge and experience to generate reasonable cartographic style sheet templates based on inspiration maps or visual art. Finally, integration of AI with cartographic design may automate part of the generalization process, a particularly promising avenue given the difficult of updating high resolution datasets and rendering new tilesets to support the 'map of everywhere'.

%One direction is to train the machine to get the knowledge about how human perceive and create the maps. The preliminary results of our case study may help users design maps with aesthetics as the model can tell which parts of maps could be further improved. Moreover, it can assist people who lack of professional knowledge such as creation of complex cartographic style sheets but are interested in creating specified types of maps or visual arts. Last but not least, it might facilitate the map generalization process as the model trained at a small scale can generalize spatial features automatically. It indicates the possibility to learn the knowledge of map generalization using advanced AI techniques.

\section*{Acknowledgments}
The authors would like to thank Bo Peng at the University of Wisconsin-Madison, Fan Zhang from the MIT Senseable city lab, and Di Zhu from the Peking University for their helpful discussions for the research.
This research was funded by the Wisconsin Alumni Research Foundation and the Trewartha Graduate Research fund.

\begin{spacing}{0.9}% tune the size by altering the parameter
      \bibliography{AI_Map_bib} % Include your own bibliography (*.bib)

\begin{thebibliography}{xx}

\bibitem[Armstrong and Xiao, 2018]{armstrong2018retrospective}
Armstrong, M.~P. and Xiao, N., 2018.
\newblock Retrospective deconstruction of statistical maps: A choropleth case
  study.
\newblock {\em Annals of the American Association of Geographers} 108(1),
  pp.~179--203.

\bibitem[Brewer and Buttenfield, 2007]{brewer2007framing}
Brewer, C.~A. and Buttenfield, B.~P., 2007.
\newblock Framing guidelines for multi-scale map design using databases at
  multiple resolutions.
\newblock {\em Cartography and Geographic Information Science} 34(1),
  pp.~3--15.

\bibitem[Buckley and Jenny, 2012]{CPcp73-buckley-jenny}
Buckley, A. and Jenny, B., 2012.
\newblock Letter from the guest editors.
\newblock {\em Cartographic Perspectives}.

\bibitem[Chiang, 2016]{chiang2016unlocking}
Chiang, Y.-Y., 2016.
\newblock Unlocking textual content from historical maps-potentials and
  applications, trends, and outlooks.
\newblock In: \emph{International Conference on Recent Trends in Image
  Processing and Pattern Recognition}, Springer, pp.~111--124.

\bibitem[Christophe and Hoarau, 2012]{christophe2012expressive}
Christophe, S. and Hoarau, C., 2012.
\newblock Expressive map design based on pop art: Revisit of semiology of
  graphics?
\newblock {\em Cartographic Perspectives} (73), pp.~61--74.

\bibitem[Christophe et al., 2016]{christophe2016map}
Christophe, S., Dum{\'e}nieu, B., Turbet, J., Hoarau, C., Mellado, N., Ory, J.,
  Loi, H., Masse, A., Arbelot, B., Vergne, R. et~al., 2016.
\newblock Map style formalization: Rendering techniques extension for
  cartography.
\newblock In: \emph{Proceedings of the Joint Symposium on Computational
  Aesthetics and Sketch Based Interfaces and Modeling and Non-Photorealistic
  Animation and Rendering}, Eurographics Association, pp.~59--68.

\bibitem[DeLucia and Black, 1987]{delucia1987comprehensive}
DeLucia, A. and Black, T., 1987.
\newblock A comprehensive approach to automatic feature generalization.
\newblock In: \emph{Proceedings of the 13th International Cartographic
  Conference}, pp.~168--191.

\bibitem[Deng et al., 2018]{deng2018like}
Deng, X., Zhu, Y. and Newsam, S., 2018.
\newblock What is it like down there? generating dense ground-level views and
  image features from overhead imagery using conditional generative adversarial
  networks.
\newblock {\em arXiv preprint arXiv:1806.05129}.

\bibitem[Duan et al., 2017]{duan2017automatic}
Duan, W., Chiang, Y.-Y., Knoblock, C.~A., Jain, V., Feldman, D., Uhl, J.~H. and
  Leyk, S., 2017.
\newblock Automatic alignment of geographic features in contemporary vector
  data and historical maps.
\newblock In: \emph{Proceedings of the 1st Workshop on Artificial Intelligence
  and Deep Learning for Geographic Knowledge Discovery}, ACM, pp.~45--54.

\bibitem[Evans et al., 2017]{evans2017livemaps}
Evans, M.~R., Mahmoody, A., Yankov, D., Teodorescu, F., Wu, W. and Berkhin, P.,
  2017.
\newblock Livemaps: Learning geo-intent from images of maps on a large scale.
\newblock In: \emph{Proceedings of the 25th ACM SIGSPATIAL International
  Conference on Advances in Geographic Information Systems}, ACM, p.~31.

\bibitem[Foerster et al., 2007]{foerster2007towards}
Foerster, T., Stoter, J. and K{\"o}bben, B., 2007.
\newblock Towards a formal classification of generalization operators.
\newblock In: \emph{Proceedings of the 23rd International Cartographic
  Conference, Moscow, Russia}, pp.~4--10.

\bibitem[Ganguli et al., 2019]{ganguli2019geogan}
Ganguli, S., Garzon, P. and Glaser, N., 2019.
\newblock Geogan: A conditional gan with reconstruction and style loss to
  generate standard layer of maps from satellite images.
\newblock {\em arXiv preprint arXiv:1902.05611}.

\bibitem[Gao et al., 2017]{gao2017designing}
Gao, S., Janowicz, K. and Zhang, D., 2017.
\newblock Designing a map legend ontology for searching map content.
\newblock {\em Advances in Ontology Design and Patterns} 32, pp.~119--130.

\bibitem[Gatys et al., 2016]{gatys2016image}
Gatys, L.~A., Ecker, A.~S. and Bethge, M., 2016.
\newblock Image style transfer using convolutional neural networks.
\newblock In: \emph{Proceedings of the IEEE Conference on Computer Vision and
  Pattern Recognition}, pp.~2414--2423.

\bibitem[Goodfellow et al., 2016]{goodfellow2016deep}
Goodfellow, I., Bengio, Y., Courville, A. and Bengio, Y., 2016.
\newblock {\em Deep learning}.
\newblock Vol.~1, MIT press Cambridge.

\bibitem[Goodfellow et al., 2014]{goodfellow2014generative}
Goodfellow, I., Pouget-Abadie, J., Mirza, M., Xu, B., Warde-Farley, D., Ozair,
  S., Courville, A. and Bengio, Y., 2014.
\newblock Generative adversarial nets.
\newblock In: \emph{Advances in neural information processing systems},
  pp.~2672--2680.

\bibitem[Hu et al., 2018]{hu2018geoai}
Hu, Y., Gao, S., Newsam, S. and Lunga, D., 2018.
\newblock {GeoAI} 2018 workshop report the 2nd acm sigspatial international
  workshop on {GeoAI}: {AI} for geographic knowledge discovery seattle, wa,
  usa-november 6, 2018.
\newblock {\em SIGSPATIAL Special} 10(3), pp.~16--16.

\bibitem[Huang et al., 2017]{huang2017densely}
Huang, G., Liu, Z., Van Der~Maaten, L. and Weinberger, K.~Q., 2017.
\newblock Densely connected convolutional networks.
\newblock In: \emph{Proceedings of the IEEE conference on computer vision and
  pattern recognition}, pp.~4700--4708.

\bibitem[Isola et al., 2017]{isola2017image}
Isola, P., Zhu, J.-Y., Zhou, T. and Efros, A.~A., 2017.
\newblock Image-to-image translation with conditional adversarial networks.
\newblock In: \emph{Proceedings of the IEEE conference on computer vision and
  pattern recognition}, pp.~1125--1134.

\bibitem[Kent and Vujakovic, 2009]{kent2009stylistic}
Kent, A.~J. and Vujakovic, P., 2009.
\newblock Stylistic diversity in european state 1: 50 000 topographic maps.
\newblock {\em The Cartographic Journal} 46(3), pp.~179--213.

\bibitem[Krizhevsky et al., 2012]{krizhevsky2012imagenet}
Krizhevsky, A., Sutskever, I. and Hinton, G.~E., 2012.
\newblock Imagenet classification with deep convolutional neural networks.
\newblock In: \emph{Advances in neural information processing systems},
  pp.~1097--1105.

\bibitem[Law et al., 2018]{law2018street}
Law, S., Seresinhe, C.~I., Shen, Y. and Gutierrez-Roig, M., 2018.
\newblock Street-frontage-net: urban image classification using deep
  convolutional neural networks.
\newblock {\em International Journal of Geographical Information Science}
  pp.~1--27.

\bibitem[LeCun et al., 2015]{lecun2015deep}
LeCun, Y., Bengio, Y. and Hinton, G., 2015.
\newblock Deep learning.
\newblock {\em nature} 521(7553), pp.~436.

\bibitem[Li and Hsu, 2018]{li2018automated}
Li, W. and Hsu, C.-Y., 2018.
\newblock Automated terrain feature identification from remote sensing imagery:
  a deep learning approach.
\newblock {\em International Journal of Geographical Information Science}
  pp.~1--24.

\bibitem[Mackaness et al., 2011]{mackaness2011generalisation}
Mackaness, W.~A., Ruas, A. and Sarjakoski, L.~T., 2011.
\newblock {\em Generalisation of geographic information: cartographic modelling
  and applications}.
\newblock Elsevier.

\bibitem[Maggiori et al., 2017]{maggiori2017convolutional}
Maggiori, E., Tarabalka, Y., Charpiat, G. and Alliez, P., 2017.
\newblock Convolutional neural networks for large-scale remote-sensing image
  classification.
\newblock {\em IEEE Transactions on Geoscience and Remote Sensing} 55(2),
  pp.~645--657.

\bibitem[Mao et al., 2017]{mao2017geoai}
Mao, H., Hu, Y., Kar, B., Gao, S. and McKenzie, G., 2017.
\newblock {GeoAI} 2017 workshop report: the 1st acm sigspatial international
  workshop on {GeoAI}:@ {AI} and deep learning for geographic knowledge
  discovery: Redondo beach, ca, usa-november 7, 2016.
\newblock {\em SIGSPATIAL Special} 9(3), pp.~25--25.

\bibitem[McMaster and Shea, 1992]{mcmaster1992generalization}
McMaster, R.~B. and Shea, K.~S., 1992.
\newblock Generalization in digital cartography.
\newblock Association of American Geographers Washington, DC.

\bibitem[Mirza and Osindero, 2014]{mirza2014conditional}
Mirza, M. and Osindero, S., 2014.
\newblock Conditional generative adversarial nets.
\newblock {\em arXiv preprint arXiv:1411.1784}.

\bibitem[Muehlenhaus, 2012]{muehlenhaus2012if}
Muehlenhaus, I., 2012.
\newblock If looks could kill: The impact of different rhetorical styles on
  persuasive geocommunication.
\newblock {\em The Cartographic Journal} 49(4), pp.~361--375.

\bibitem[Peterson, 2011]{peterson2011travels}
Peterson, M.~P., 2011.
\newblock Travels with ipad maps.
\newblock {\em Cartographic Perspectives} (68), pp.~75--82.

\bibitem[Peterson, 2014]{shook2014mapping}
Peterson, M.~P., 2014.
\newblock Mapping in the cloud.

\bibitem[Raposo, 2017]{rap}
Raposo, P., 2017.
\newblock Scale and generalization.
\newblock In: \emph{The Geographic Information Science \& Technology Body of
  Knowledge (4th Quarter 2017 Edition)}, John P. Wilson.

\bibitem[Reed et al., 2016]{reed2016generative}
Reed, S., Akata, Z., Yan, X., Logeswaran, L., Schiele, B. and Lee, H., 2016.
\newblock Generative adversarial text to image synthesis.
\newblock {\em arXiv preprint arXiv:1605.05396}.

\bibitem[Regnauld and McMaster, 2007]{regnauld2007synoptic}
Regnauld, N. and McMaster, R.~B., 2007.
\newblock A synoptic view of generalisation operators.
\newblock In: \emph{Generalisation of geographic information}, Elsevier,
  pp.~37--66.

\bibitem[Roth, forthcoming]{rothforth}
Roth, R., forthcoming.
\newblock Cartographic design as visual storytelling: synthesis \& review of
  map-based narratives, genres, and tropes.

\bibitem[Roth et al., 2011]{roth2011typology}
Roth, R.~E., Brewer, C.~A. and Stryker, M.~S., 2011.
\newblock A typology of operators for maintaining legible map designs at
  multiple scales.
\newblock {\em Cartographic Perspectives} (68), pp.~29--64.

\bibitem[Roth et al., 2015]{CPcp78-roth-et-al}
Roth, R.~E., Donohue, R.~G., Sack, C.~M., Wallace, T.~R. and Buckingham, T.,
  2015.
\newblock A process for keeping pace with evolving web mapping technologies.
\newblock {\em Cartographic Perspectives} (78), pp.~25--52.

\bibitem[Shen et al., 2018]{shen2018new}
Shen, Y., Ai, T., Wang, L. and Zhou, J., 2018.
\newblock A new approach to simplifying polygonal and linear features using
  superpixel segmentation.
\newblock {\em International Journal of Geographical Information Science}
  32(10), pp.~2023--2054.

\bibitem[Srivastava et al., 2018]{srivastava2018multilabel}
Srivastava, S., Vargas-Mu{\~n}oz, J.~E., Swinkels, D. and Tuia, D., 2018.
\newblock Multilabel building functions classification from ground pictures
  using convolutional neural networks.
\newblock In: \emph{Proceedings of the 2nd ACM SIGSPATIAL International
  Workshop on AI for Geographic Knowledge Discovery}, ACM, pp.~43--46.

\bibitem[Stanislawski et al., 2014]{stanislawski2014generalisation}
Stanislawski, L.~V., Buttenfield, B.~P., Bereuter, P., Savino, S. and Brewer,
  C.~A., 2014.
\newblock Generalisation operators.
\newblock In: \emph{Abstracting Geographic Information in a Data Rich World},
  Springer, pp.~157--195.

\bibitem[Stoter, 2005]{stoter2005generalisation}
Stoter, J., 2005.
\newblock Generalisation within nma’s in the 21st century.
\newblock In: \emph{Proceedings of the International Cartographic Conference}.

\bibitem[Szegedy et al., 2016]{szegedy2016rethinking}
Szegedy, C., Vanhoucke, V., Ioffe, S., Shlens, J. and Wojna, Z., 2016.
\newblock Rethinking the inception architecture for computer vision.
\newblock In: \emph{Proceedings of the IEEE conference on computer vision and
  pattern recognition}, pp.~2818--2826.

\bibitem[VoPham et al., 2018]{vopham2018emerging}
VoPham, T., Hart, J.~E., Laden, F. and Chiang, Y.-Y., 2018.
\newblock Emerging trends in geospatial artificial intelligence ({geoAI}):
  potential applications for environmental epidemiology.
\newblock {\em Environmental Health} 17(1), pp.~40.

\bibitem[Xu and Zhao, 2018]{xu2018satellite}
Xu, C. and Zhao, B., 2018.
\newblock Satellite image spoofing: Creating remote sensing dataset with
  generative adversarial networks.
\newblock In: \emph{10th International Conference on Geographic Information
  Science (GIScience 2018)}, Schloss Dagstuhl-Leibniz-Zentrum fuer Informatik.

\bibitem[Zhang et al., 2019]{zhang2019social}
Zhang, F., Wu, L., Zhu, D. and Liu, Y., 2019.
\newblock Social sensing from street-level imagery: A case study in learning
  spatio-temporal urban mobility patterns.
\newblock {\em ISPRS Journal of Photogrammetry and Remote Sensing} 153,
  pp.~48--58.

\bibitem[Zhang et al., 2018]{zhang2018measuring}
Zhang, F., Zhou, B., Liu, L., Liu, Y., Fung, H.~H., Lin, H. and Ratti, C.,
  2018.
\newblock Measuring human perceptions of a large-scale urban region using
  machine learning.
\newblock {\em Landscape and Urban Planning} 180, pp.~148--160.

\bibitem[Zhou et al., 2018]{zhou2018deep}
Zhou, X., Li, W., Arundel, S.~T. and Liu, J., 2018.
\newblock Deep convolutional neural networks for map-type classification.
\newblock {\em arXiv preprint arXiv:1805.10402}.

\bibitem[Zhu et al., 2019]{zhu2019spatial}
Zhu, D., Cheng, X., Zhang, F., Yao, X., Gao, Y. and Liu, Y., 2019.
\newblock Spatial interpolation using conditional generative adversarial neural
  networks.
\newblock {\em International Journal of Geographical Information Science}
  pp.~1--24.

\bibitem[Zhu et al., 2017]{CycleGAN2017}
Zhu, J.-Y., Park, T., Isola, P. and Efros, A.~A., 2017.
\newblock Unpaired image-to-image translation using cycle-consistent
  adversarial networks.
\newblock In: \emph{Proceedings of the IEEE international conference on
  computer vision}, pp.~2223--2232.

\bibitem[Zou et al., 2015]{zou2015deep}
Zou, Q., Ni, L., Zhang, T. and Wang, Q., 2015.
\newblock Deep learning based feature selection for remote sensing scene
  classification.
\newblock {\em IEEE Geoscience and Remote Sensing Letters} 12(11),
  pp.~2321--2325.

\end{thebibliography}
\end{spacing}

\label{LastPage} % added 13-03-2018 by Juergen Bierwirth to get the \pageref{LastPage} result (page numbering in the header)
\end{document}